\documentclass[preprint,12pt,times,3p]{elsarticle}

\usepackage{amssymb}

\usepackage{lineno}

\usepackage{graphicx}		
\usepackage{amsmath}
\usepackage{amssymb}
\usepackage{xcolor}         
\usepackage{url}
\usepackage{booktabs}
\usepackage{layouts}
\usepackage{subcaption}

\usepackage[noend]{algpseudocode}
\usepackage{algorithm}

\usepackage{tikz}
\usetikzlibrary{decorations.pathreplacing, positioning, shapes, arrows.meta, intersections}
\tikzset{>={Latex[length=1.5mm, width=1mm]}}

\usepackage{hyperref}
\hypersetup{
    pdfborder={0 0 0},
    colorlinks=true,
    bookmarks=true,
    bookmarksnumbered=true
}

\journal{Spatial Statistics}

\newcommand{\x}{\boldsymbol{x}}

\newcommand{\n}{\boldsymbol{n}}

\renewcommand{\n}{\boldsymbol{n}}
\newcommand{\W}{\boldsymbol{W}}
\renewcommand{\b}{\boldsymbol{b}}

\newcommand{\thetabf}{\boldsymbol{\theta}}
\newcommand{\phibf}{\boldsymbol{\phi}}

\makeatletter
\newcommand\thefontsize[1]{{#1: \f@size pt\par}}
\makeatother

\begin{document}

\begin{frontmatter}



\title{Bayesian Physics Informed Neural Networks for Data Assimilation and Spatio-Temporal Modelling of Wildfires}


\author[csiro]{Joel Janek Dabrowski\corref{cor1}}
\ead{Joel.Dabrowski@data61.csiro.au}
\author[csiro]{Daniel Edward Pagendam}
\author[csiro]{James Hilton}
\author[csiro]{Conrad Sanderson}
\author[csiro]{Daniel MacKinlay}
\author[csiro]{Carolyn Huston}
\author[csiro]{Andrew Bolt}
\author[csiro]{Petra Kuhnert}

\affiliation[csiro]{organization={Data61 / CSIRO},
            country={Australia}}
\cortext[cor1]{Corresponding author}

\begin{abstract}
    We apply the Physics Informed Neural Network (PINN) to the problem of wildfire fire-front modelling. We use the PINN to solve the level-set equation, which is a partial differential equation that models a fire-front through the zero-level-set of a level-set function. The result is a PINN that simulates a fire-front as it propagates through the spatio-temporal domain. We show that popular optimisation cost functions used in the literature can result in PINNs that fail to maintain temporal continuity in modelled fire-fronts when there are extreme changes in exogenous forcing variables such as wind direction. We thus propose novel additions to the optimisation cost function that improves temporal continuity under these extreme changes. Furthermore, we develop an approach to perform data assimilation within the PINN such that the PINN predictions are drawn towards observations of the fire-front. Finally, we incorporate our novel approaches into a Bayesian PINN (B-PINN) to provide uncertainty quantification in the fire-front predictions. This is significant as the standard solver, the level-set method, does not naturally offer the capability for data assimilation and uncertainty quantification. Our results show that, with our novel approaches, the B-PINN can produce accurate predictions with high quality uncertainty quantification on real-world data.
\end{abstract}

%


\end{frontmatter}


\section{Introduction}

The wildfires that swept Australia in 2019 and 2020 burned over 19 million hectares of land, destroyed over 3094 homes, killed an estimated 1.25 billion animals, and resulted in the loss of 33 human lives \cite{wwf2022australia}. This is one example of the disastrous effects of wildfires that occur across the globe every year \cite{Bowman_2020}. Capturing wildfire dynamics in a model to allow emergency management decision-makers to readily access, interpret and make trusted decisions in the face of uncertainty is paramount as highlighted by \cite{huston2022creating}. This is a challenging modelling task given the complexity of a wildfire and the uncertainty associated with its driving forces.

The level-set method is a common approach used to model wildfires and is the basis of many wildfire prediction platforms such as SPARK \cite{miller2015spark} and WRF-SFIRE \cite{mandel2011coupled}. In this method, a fire-front is modelled in the form of a Partial Differential Equation (PDE) called the level-set equation. The level-set method has the advantage that it is able to follow changes in the topology, such as shape splits and mergers \cite{Osher2001Level}. However, it requires complicated finite difference derivative approximations over finely-scaled discretised grids and it also requires a reinitialisation scheme to avoid numerical errors caused by large variations in the modelled derivatives \cite{Osher2001Level,sethian1999level}. 

Especially in the context of wildfires, a key disadvantage of the level-set method is that it does not provide any natural means for uncertainty quantification and data assimilation \cite{yoo2022bayesian,dabrowski2022towards}. Given the uncertainty around the driving forces of a fire (such as wind, fuel type, fuel load, etc.), there will be an inherent uncertainty around any predictions of the location and the dynamics of a fire-front. In the absence of uncertainty quantification, making informed decisions regarding a wildfire can be challenging \cite{huston2022creating,kuhnert2018making}. Furthermore, without data assimilation, inaccuracies can grow into large errors with no means for correction \cite{Rochoux2013Regional}, which also result in increased uncertainty in the predictions over time.

The Physics Informed Neural Network (PINN) \cite{Lagaris1998Artificial,raissi2019physics} is a machine learning approach that uses a neural network to solve a PDE. PINNs have recently received increased interest as they take advantage of the non-linearity, differentiability, and the universal approximation properties of neural networks to provide an approximate solution to PDEs in a functional form over \textit{continuous} time and space \cite{Cuomo2022Scientific}. As such, a trained PINN can provide predictions at various spatio-temporal resolutions without the need for retraining \cite{markidis2021old}. Furthermore, the PINN offers a means to perform data assimilation (e.g., \cite{he2020physics}) and also provide uncertainty quantification via the Bayesian PINN (B-PINN) \cite{yang2021bpinns}. In this study, we take advantage of these properties and explore the application of the PINN to solve the level-set equation for wildfire fire-front modelling. However, in our application we encounter the challenge where the PINN (in its standard form) can fail to maintain temporal continuity in its solution \cite{wang2022respecting,mojgani2022lagrangian}. A key objective of this study is to address this challenge.

Our primary contributions are (i) we propose a novel approach which leads to  PINN-produced fire-fronts that maintain temporal continuity under non-linear changes to exogenous forcing variables (such as wind and fuel load) and  demonstrate this on a synthetic dataset; (ii) we propose an approach to perform data assimilation in the PINN for the level-set method; and (iii) we demonstrate data assimilation and uncertainty quantification with a PINN on real-world fire data, which is significant given that the level-set method does not naturally offer these capabilities. In our results, we treat the level-set method as the gold standard and show that the PINN is able to produce similar results to the level-set method, but with the added benefits of data assimilation and uncertainty quantification.

\section{Related Work}
\label{sec:relatedWork}

Spatio-temporal models have been widely used to capture complex interactions and dynamics of environmental systems.  Approaches for modelling have included traditional machine learning approaches such as random forests \cite{kuhnert2010incorporating}; statistical approaches such as kriging \cite{ZammitMangion2021frk}; more sophisticated physical statistical models \cite{kuhnert2014physical} using Bayesian Hierarchical modelling frameworks \cite{wikle1998hierarchical,wikle2010general,gladish2016spatiotemporal,wikle2019spatio}; and more recently, emulation approaches that use deep learning to approximate physical systems \cite{bolt2022spatiotemporal}. \citet{bolt2022spatiotemporal} in particular has demonstrated machine learning-based emulation approaches for the SPARK wildfire model. 

The spatio-temporal modelling of wildfires is commonly performed using the level-set method  \cite{mallet2009modeling,hilton2015effects,hilton2016curvature,miller2015spark,mandel2011coupled,yoo2022bayesian}. This method uses finite difference approximations of the level-set equation, such as the hyperbolic upwind techniques presented in this study \cite{osher2003level,sethian1999level}. These approximations can be complicated and usually require a fine spatio-temporal grid to maintain stability; which can be computationally expensive, especially when modelling large regions. Without fine grids, small errors caused by inaccuracies in the linear approximations of derivatives or aliasing can grow into large artefacts resulting in a diverging solution. Furthermore, as the level-set function evolves over time, it may evolve into a form that has high variations in its derivatives which can introduce significant errors in the discrete approximations. This is typically controlled by periodically reinitialising the level-set function \cite{osher2003level,sethian1999level}. Achieving an accurate reinitialisation is however challenging \cite{sethian1999level}.

Another disadvantage of the level-set method is that it does not offer any natural means for assimilating data into predictions over time. Data assimilation is often performed using Bayesian filtering methods \cite{Srivas2016Wildfire,Xue2012Data,Silva2014Application}. It has been shown that it is possible to incorporate the level-set method into the Bayesian filtering paradigm, such as with the Best Linear Unbiased Estimator (BLUE) \cite{Rochoux2013Regional} and the particle filter \cite{dabrowski2022towards}. However, these approaches can add a significant amount of computational complexity to an already complex approach, and it can be challenging to ensure that the approaches operate within the constraints of the level-set method (e.g., maintaining smooth gradients in the level-set function and performing reinitialisation).

The final disadvantage of the level-set method in the context of wildfire modelling, is that it does not offer any natural means for uncertainty quantification. Representing uncertainty requires formulating the level-set function (and its derivatives) within a probabilistic modelling framework and performing the finite difference numerical approximations within this paradigm. One approach is to combine a mechanistic dynamic level-set model and a stochastic spatio-temporal dynamic front velocity model using a Bayesian hierarchical model \cite{yoo2022bayesian}. Another approach is to integrate the level-set method into a particle filter \cite{dabrowski2022towards}. Both of these approaches involve inference methods that can be computationally expensive and compound with the already intensive approach of the level-set method. Finally, in one other notable approach, an echo state network is incorporated into the level-set method to model non-linear fire-front speeds in the normal direction \cite{yoo2023using}. A convenient by-product of the approach is that it provides a means for calibrated uncertainty quantification.

The PINN offers an alternative approach to solving PDEs without the need for numerical discretisation and finite differences. Instead, the PINN makes use of automatic differentiation (also known as algorithmic differentiation) \cite{baydin2018automatic} and provides an approximate PDE solution as the output of neural network. Furthermore, the PINN is able to solve both the forward problem (solving the PDE) and the inverse problem (discovery of the PDE) \cite{raissi2019physics}. With this and recent technological advancements (such as optimisation frameworks and automatic differentiation), PINNs have been recently applied to a wide variety of applications and several variations of PINNs have been proposed. This is especially exhibited in several recent surveys on PINNs \cite{Cuomo2022Scientific,Karniadakis2021Physics,Kollmannsberger2021Physics}.

One particular variation of the PINN that is of interest to this study is the Bayesian PINN (B-PINN) \cite{yang2021bpinns} as it offers a means to quantify uncertainty. Given that PINNs are often applied to problems where there is a lack of data, uncertainty is best represented in the form of epistemic uncertainty \cite{hullermeier2021aleatoric}. The Bayesian statistical paradigm offers epistemic uncertainty quantification by defining the uncertainty in the terms of the PINN model parameters. In the B-PINN, the posterior distribution of the PINN parameters given the data is computed using methods such as variational inference or Markov Chain Monte Carlo (MCMC). Predictions with uncertainty quantification can then be provided using the posterior predictive distribution. Other approaches to incorporating uncertainty in the PINN include randomised prior functions \cite{Costabal2020Physics}, Monte Carlo Dropout \cite{zhang2019quantifying}, and adversarial approaches \cite{yang2019adversarial,Gao2022Wasserstein}. Variational inference is however the most commonly used approach.

PINNs are however notoriously difficult to optimise as they may have long training times and can potentially have limited accuracy \cite{markidis2021old,krishnapriyan2021characterizing}. The universal approximation theorem \cite{hornik1989multilayer} dictates that neural networks are capable of learning a very broad class of functions with high accuracy given sufficient data and network capacity. However, it is often not a lack in capacity of the neural network that limits a network's accuracy, but rather challenges associated with optimising over a complex landscape \cite{krishnapriyan2021characterizing}. These challenges include gradient pathologies \cite{wang2021understanding} and spectral bias \cite{wang2021eigenvector,wang2022when} (where a neural network learns low frequency features first and may require an extensive number training epochs to learn higher frequency components).

Another significant challenge with the PINN is that it can fail to maintain temporal continuity\footnote{\citet{wang2022respecting} refers to the temporal continuity problem as a ``causality'' problem. We do not use this term given its definitions in various other contexts.} in its solution \cite{wang2022respecting,mojgani2022lagrangian}. PINN's trained using mean squared error loss functions that are prevalent in the literature, tend to result in solutions that violate the physical laws that govern the temporal progression of a system in continuous-time; we observe this particularly where there are abrupt changes in exogenous forcing variables such as wind direction. Several strategies have been proposed to address such problems, including sequential sampling \cite{krishnapriyan2021characterizing,mattey2022novel}, sequential weighting \cite{wang2022respecting}, and reformulating PINNs on a Lagrangian frame of reference \cite{mojgani2022lagrangian}. Sequential strategies typically force the PINN to train in a sequential manner, but they do not enforce temporal continuity of the solutions. The discrete PINN \cite{raissi2019physics} offers another possible solution as it solves the PDE sequentially. However, with temporally varying, exogenous forcing variables (such as wind vector fields), the neural network requires retraining at each time-step. The Recurrent Neural Network (RNN) provides an alternative to the discrete PINN where Runge-Kutta integration is encoded in an RNN cell \cite{nascimento2020tutorial,yucesan2020physics}. Retraining is not necessary, but labelled data are required at each time-step. Such data is typically not available in wildfire applications.

At the time of writing, we are only aware of two applications of PINNs to the level-set equation. In the first study, \citet{zubov2021neuralpde} provide a simplified example of implementing the level-set equation with a PINN. However all the results and discussion relate to convergence of the PINN and do not provide any insight into the solution that the PINN obtains. In the second study, \citet{bottero2021physics} provide a technical report on applying the PINN to modelling wildfires with the level-set equation. The approach implements the level-set equation and also includes other atmospheric modelling PDEs from the WRF-SFIRE simulator \cite{mandel2011coupled}. However, the focus is on the implementation in the Julia language rather than on more broader challenges associated with the application of a PINN to the level-set equation and wildfires.  Furthermore, a key limitation of the implementation in \cite{mandel2011coupled} is that it is constrained to a static fuel map, which severely limits their approach to static conditions. Finally, neither of the two studies consider data assimilation and uncertainty quantification.

\section{Methods}

\subsection{The Level-Set Equation and the Level-Set Method}

\subsubsection{The Level-Set Equation}
\label{sec:levelSetEquation}

The \textit{level-set method} implicitly models a moving fire-front in two-dimensional space using an auxiliary three-dimensional surface called the \textit{level-set function}. This function is evolved over space and time using a Partial Differential Equation (PDE) known as the \textit{level-set equation}. The moving front is represented by the \textit{zero-level-set}, which comprises a set of closed curves or isochrones located where the level-set function equates to zero. Let $u(t,x,y)$ denote the level-set function where $t$ denotes time and $(x,y)$ denotes two-dimensional space\footnote{For notational convenience, we may drop the dependence on $(t,x,y)$ in functions and vector fields. For example we may represent $u(t,x,y)$ in shorthand form as $u$.}.  The level-set equation is given by \cite{Osher2001Level,sethian1999level}
\begin{align}
    \label{eq:levelSetEquation1}
    \frac{\partial u}{\partial t} + s \| \nabla u \| = 0,
\end{align}
where $s(t,x,y) \geq 0$ is a scalar function describing the speed at which the level-set function moves in its normal direction, $\| \cdot \|$ indicates the Euclidean norm, and $\nabla$ is the vector differential operator over space $(x,y)$ such that 
\begin{align}
    \nabla u = \left[ \frac{\partial u}{\partial x}, \frac{\partial u}{\partial y} \right]^\top .
\end{align}

In the context of a wildfire, $s$ may specify how the fire expands over time (in the normal direction) and includes factors such as fuel properties and landscape topography. In the simplest possible scenario, $s$ is not a result of any exogenous forcing, it can be viewed as an endogenous vector field that causes the level-set function to propagate outwards in the normal direction with a velocity proportional to its spatial gradient \cite{osher2003level}. That is, let $\hat{\n}(t,x,y)$ be the normal vector of the level-set function at time $t$ and location $(x,y)$ such that the internal vector field\footnote{Vector fields are denoted by upper-case letters with arrows and vectors are denoted by boldface lower-case letters.} is given by
\begin{align}
    \vec{S}(t,x,y) = s(t,x,y) \hat{\n}(t,x,y).
\end{align}
In the more realistic situation where a fire is driven by wind, we can consider an external vector field 
\begin{align}
    \vec{W}(t,x,y) = [w_x(t,x,y), w_y(t,x,y)]^\top ,
\end{align}
that advects the level set function across space. To include $\vec{W}$, the level-set equation can more generally be represented in the form of an advection equation with a resultant (the superposition of endogenous and exogenous forcings) vector field  $\vec{C}$ as follows \cite{osher2003level}
\begin{align}
    \label{eq:levelSetEquation}
    \frac{\partial u}{\partial t} + \vec{C} \cdot \nabla u = 0.
\end{align}
According to the Rothermal wildfire Rate Of Spread (ROS) model \cite{rothermel1972mathematical}, the vector field $\vec{C}$ can be expressed as
\begin{align}
    \label{eq:rothermal}
    \vec{C} = \vec{S} + \vec{W}.
\end{align}
With this, we have that 
\begin{align}
    \frac{\partial u}{\partial t} + (s \hat{\n} + \vec{W}) \cdot \nabla u, &= 0, \nonumber \\
    \frac{\partial u}{\partial t} + s \| \nabla u \| + \vec{W} \cdot \nabla u, &= 0.
\end{align}
Here the second term corresponds to the level-set equation of (\ref{eq:levelSetEquation1}) and causes the fire to expand in the direction of the normal vectors at each point on the fire-front. The third term corresponds to an advection equation and causes the fire to shift through space as a result of the wind vector at each point on the fire-front. This term can however cause the fire to shift in such a way that it backtracks on itself if $\vec{W} \cdot \hat{\n} < 0$. As this is typically not physically possible, the condition of $\vec{W} \cdot \hat{\n} \geq 0$ is enforced as follows \cite{hilton2015effects}:
\begin{align}
    \vec{W} \cdot \hat{\n} =
    \begin{cases}
        \vec{W} \cdot \hat{\n}, & \vec{W} \cdot \hat{\n} > 0, \\
        0, & \vec{W} \cdot \hat{\n} \leq 0.
    \end{cases}
\end{align}

The fire-front at time $t$ is represented by the zero-level-set $\Gamma_t$ of the level-set function $u$, which is a closed curve given by
\begin{align}
    \Gamma_t = \left\{ (x,y) | u(t,x,y) = 0 \right\}.
\end{align}
As the level-set function evolves over time and space, the zero-level-set can be tracked to provide a simulation of a wildfire fire-front.

\subsubsection{The Level-Set Function and Distance Functions}
\label{sec:signedDistanceFunction}

The level-set function $u$ is required to be initialised to some three-dimensional form at time $t=0$. There are a wide range of initial level-set function forms to select from. The main property of the level-set function is that it is positive on the exterior, negative on the interior, and zero on the boundary of the zero-level-set \cite{osher2003level}. This property ensures that the zero-level-set exists. Smoothness with small variations in the gradient $\nabla u$ is also a desirable property for methods that rely on numerical approximations (such as the level-set method), as it reduces numerical errors. Heavyside functions and signed distance functions have these properties and are thus common choices for the level-set function. Signed distance functions have been shown to provide more accurate numerical solutions given the two additional properties where: (i) $u$ provides the shortest distance from $(x,y)$ to the zero-level-set, such that, for distance function $d(x,y)$,
\begin{align}
    u(t=0, x,y) =
    \begin{cases}
        -d(x,y), & \text{$\forall$ $(x,y)$ interior to $\Gamma_{t = 0}$}, \\
        0, & \forall ~ (x,y) \in \Gamma_{t = 0} ,\\
        d(x,y), & \text{$\forall$ $(x,y)$ exterior to $\Gamma_{t = 0}$}.
    \end{cases}
\end{align}
and (ii) 
\begin{align}
    \| \nabla u \| = 1,  \quad \forall (x,y).
\end{align}
These properties ensure smoothness and small variations in the gradients $\nabla u$ which assist in reducing errors in the numerical approximations of derivatives. 

The level-set function may be initialised such that the zero-level set is constrained to some initial observed fire-front. This can be achieved by calculating the signed-distance from a point or polygon representation of the observed fire-front. The fires considered in this study begin from a point source and we thus initialise the fires with either a cone or elliptical cone. 

The cone signed-distance function is given by
\begin{align}
    \label{eq:cone}
    u(t=0,x,y) = \sqrt{x^2 + y^2} - r,
\end{align}
where $r$ is a constant. In this form, the zero-level-set is a circle radius $r$. Similarly, for an elliptical zero-level-set, an elliptical cone is given by
\begin{align}
    \label{eq:ellipticalCone}
    u(t=0,x,y) = \sqrt{ \frac{ \left( x \cos(\alpha) + y \sin(\alpha) \right)^2 }{a^2} + \frac{ \left( x \sin(\alpha) - y \cos(\alpha) \right)^2 }{b^2} } - r,
\end{align}
where $a$ and $b$ define the width and height of the ellipse, $\alpha$ is the rotation angle of the ellipse, and $r$ defines the offset of the cone from the zero-$xy$ plane. The elliptical cone does not necessarily conform to the signed-distance function condition of $\| \nabla u \| = 1$, however we often found it to be more appropriate initial form for $u$ given that fire-fronts often have an elliptical shape \cite{green1983fire}.

\subsubsection{Discretisation}

The level-set method solves the level-set equation using accurate finite difference numerical methods. For this, space and time are discretised into a grid. Let $\Delta t$ denote the discrete time-step size, let $\Delta x$ denote the discrete step size in $x$, and let $\Delta y$ denote the discrete step size in $y$. Furthermore, let $n$ denote the discrete index for time and let $(i,j)$ denote the discrete indexes for space such that $u(n,i,j)$ denotes the level-set function evaluated at the point in the spatio-temporal grid indexed by $n$, $i$, and $j$.

Under spatial discretisation of $u$, the forward and backwards difference operators in the $x$ and $y-$directions are defined by \cite{sethian1999level}
\begin{align}
    \label{eq:diffpx}
    D^{+x} u(n,i,j) & = \frac{u(n,i + 1,j) - u(n,i,j)}{\Delta x}, \\
    \label{eq:diffnx}
    D^{-x} u(n,i,j) & = \frac{u(n,i,j) - u(n,i - 1,j)}{\Delta x}, \\
    \label{eq:diffpy}
    D^{+y} u(n,i,j) & = \frac{u(n,i,j + 1) - u(n,i,j)}{\Delta y}, \quad \textup{and}\\
    \label{eq:diffny}
    D^{-y} u(n,i,j) & = \frac{u(n,i,j) - u(n,i,j - 1)}{\Delta y}.
\end{align}
The positive exponent in the difference operator denotes forwards differencing and the negative exponent in the difference operator denotes backwards differencing. With this, the first-order hyperbolic upwind finite-difference method approximation to (\ref{eq:levelSetEquation}) is given by \cite{sethian1999level}
\begin{align}
    \label{eq:discrete_level_set}
    u(n+1,i,j) = u(n,i,j) - \Delta t \left( \max(\vec{C} \cdot \hat{\n}, 0) \nabla_{ij}^{+} + \min(\vec{C} \cdot \hat{\n},0) \nabla_{ij}^{-} \right),
\end{align}
where
\begin{align}
    \label{eq:nabla_plus}
    \nabla_{ij}^{+} = \Big[& \max(D^{-x} u(n,i,j), 0)^2 + \min(D^{+x} u(n,i,j), 0)^2 + \nonumber \\
    & \max(D^{-y} u(n,i,j), 0)^2 + \min(D^{+y} u(n,i,j), 0)^2 \Big]^{1/2},
\end{align}
and 
\begin{align}
    \label{eq:nabla_minus}
    \nabla_{ij}^{-} = \Big[& \min(D^{-x} u(n,i,j), 0)^2 + \max(D^{+x} u(n,i,j), 0)^2 + \nonumber \\
    & \min(D^{-y} u(n,i,j), 0)^2 + \max(D^{+y} u(n,i,j), 0)^2 \Big]^{1/2}.
\end{align}
Equation (\ref{eq:discrete_level_set}) is evolved from time index $n$ to time index $n+1$ by updating $u(n,i,j)$ at each point $(i,j)$ in the spatial grid to produce the prediction $u(n+1,i,j)$ at time $n+1$.

\subsubsection{Reinitialisation}

The level-set function will often not maintain a signed distance function form as it is evolved over time under $\vec{C}$ and numerical inaccuracies may occur if there are large variations in the gradients $\nabla u$. The level-set function can be reinitialised to a signed distance function by finding a new $u$ with the same zero-level-set but with $\| \nabla u \| = 1$ $\forall (x,y)$. Given that we are only interested in the zero-level-set, the approach is to modify all other level-sets in a way that forces $u$ to conform to the signed distance function properties. Various approaches have been proposed to achieve this, one of which is to use the following partial differential equation \cite{sussman1994level}:
\begin{align}
    \label{eq:reinitialisation}
    \frac{\partial u}{\partial t} + \text{sign}(u_0)( \| \nabla u \| - 1) = 0.
\end{align}
Given an initial condition for $u_0$ (which is the level-set function with the desired zero-level-set), solving (\ref{eq:reinitialisation}) to steady state provides a new $u$ with the property $\|\nabla  u \| = 1$ and maintains the zero-level-set of the initial $u_0$ \cite{sussman1994level,sethian1999level}. Intuitively, the $\text{sign}(u_0)$ factor enforces the preservation of the $u_0$ zero-level-set and the $( \| \nabla u \| - 1)$ factor enforces the property $\| \nabla u \| = 1$. Equation (\ref{eq:reinitialisation}) can be discretised in a similar manner to the level-set equation and the discontinuity in the sign function can be numerically smeared out using \cite{sussman1994level,osher2003level}
\begin{align}
     \text{sign}(u_0) \approx \frac{u_0}{\sqrt{u_0^2 + (\Delta x)^2}}.
\end{align}
The reinitialisation scheme involves pausing the time evolution of $u$ every $T_r$ discrete time-steps, reinitialising $u$, and then resuming the time evolution with the reinitialised $u$. The reinitialisation is performed by solving (\ref{eq:reinitialisation}) for a predefined number of discrete steps with $u_0$ set as the last state of $u$ before the reinitialisation scheme is run. 

Owing to numerical approximations, the zero-level-set can shift considerably during reinitialisation and it is thus recommended that reinitialisation should be used sparingly \cite{sethian1999level}. In general, reinitialisation is less necessary if gradient $\nabla u$ does not vary significantly.

\subsubsection{Algorithm}

The level-set method algorithm begins with initialising the level-set function over the spatial grid at discrete time index $n=0$ (e.g., as a signed distance function). At each discrete time step, the level-set function is updated at each point in the spatial grid according to (\ref{eq:discrete_level_set}). Every $T_r$ discrete time-steps the level-set function is reinitialised according to (\ref{eq:reinitialisation}). The algorithm for the level-set method is presented in Algorithm \ref{alg:level_set_method}.

\begin{algorithm}[!t]
        \begin{algorithmic}[1]
            \Require the discrete time step size $\Delta t$, the spatial grid size $(\Delta x, \Delta y)$, the total number of time steps $T$, and the reinitialisation time $T_r$.
            \State Initialise the level-set function over the spatial grid for time step $n=0$.
            \For{$n \in (1,T)$}
            \State Calculate the forward and backward differences over $x$ and $y$ using (\ref{eq:diffpx}) through (\ref{eq:diffny}).
            \State Calculate $\nabla_{ij}^{+}$ and $\nabla_{ij}^{-}$ according to (\ref{eq:nabla_plus}) and (\ref{eq:nabla_minus}) respectively.
            \State Update the level-set function according to (\ref{eq:discrete_level_set}).
            \If{$n \mod T_r = 0$}
            \State Perform a reinitialisation step according to (\ref{eq:reinitialisation}).
            \EndIf
            \EndFor
        \end{algorithmic}
    \caption{Level-set method algorithm.}
    \label{alg:level_set_method}
\end{algorithm}
\subsection{Physics Informed Neural Network}

\subsubsection{The PINN Architecture}

The PINN approximates the solution to a PDE by representing the solution in a functional form with a neural network. In this study, the PINN is applied to solve the level-set equation given in (\ref{eq:levelSetEquation}) such that $u(t, x, y)$ is approximated with a neural network $\tilde{u}(t, x, y, s, w_x, w_y)$. The level-set function, $u$, evolves through time as a function of the forcings, $\vec{C}$, according to (\ref{eq:levelSetEquation}), and consequently $\vec{C} = s \hat{\n} + [w_x, w_y]^T$ are provided as inputs to neural network $\tilde{u}$ so that the model can respond to changes in the forcings over time. 

The PINNs considered in this study make use of the two-layered fully-connected Feed-forward Neural Network (FNN) architecture, which is given by
\begin{align}
    \label{eq:mlp}
    \tilde{u}(t, x, y, s, w_x, w_y; \thetabf) = \W_3 g(\W_2 g(\W_1 [t, x, y, s, w_x, w_y]^\top + \b_1 ) + \b_2) + \b_3,
\end{align}
where $g$ is the activation function (e.g., $\tanh$ or the Rectified Linear Unit (ReLU)); $\W_1$, $\W_2$, and $\W_3$ are weight matrices; $\b_1$, $\b_2$, and $\b_3$ are bias vectors; and $[t, x, y, s, w_x, w_y]^\top$ is an input vector. The weight matrices and bias vectors are the tunable parameters of the neural network (and the PINN) and are denoted by $\thetabf = \{ \W_1, \W_2, \W_3, \b_1, \b_2, \b_3 \}$.

The function in (\ref{eq:mlp}) provides a solution to the PDE in functional form and this solution is thus provided over the continuous spatio-temporal domain. Furthermore, (\ref{eq:mlp}) is differentiable with respect to $t$, $x$, and $y$ and these derivatives can be calculated using automatic differentiation \cite{baydin2018automatic}. Using these derivatives, the PINN can be constructed in terms of the level-set equation in (\ref{eq:levelSetEquation}) (discussed in  section \ref{sec:levelSetEquation}) as follows \cite{raissi2019physics, raissi2017physics}
\begin{align}
    \label{eq:pinn}
    f(t,x,y,s,w_x,w_y; \thetabf) := \frac{\partial \tilde{u}}{\partial t} + \vec{C} \cdot \nabla \tilde{u}.
\end{align}
Note that $\thetabf$ is the same set of parameters for both the neural network and the PINN. 

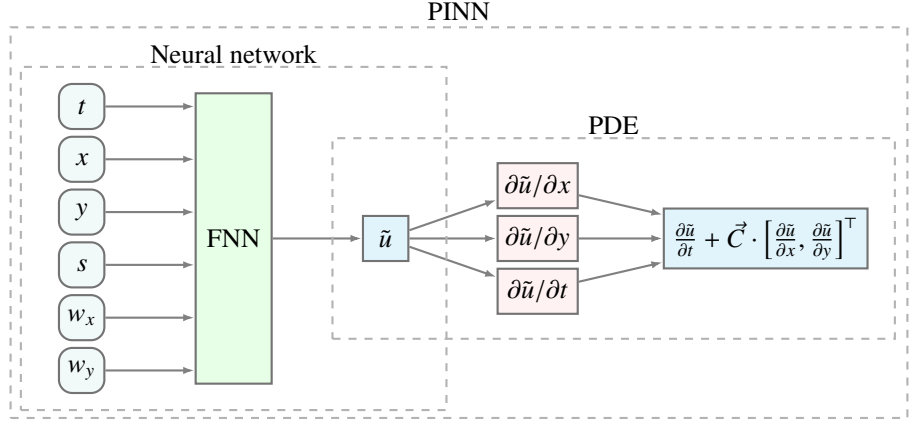
\begin{figure}[!t]
    \centering

\def\horisep{2.0cm}
\def\vertsep{0.7cm}

\begin{tikzpicture}[->,draw=black!50, thick]
    \begin{footnotesize}
        \tikzstyle{node}=[rectangle, rounded corners=5pt, draw=black!55, fill=teal!5, inner sep=1pt, minimum height=17pt, minimum width=17pt]
        \tikzstyle{node1}=[rectangle,draw=black!55, fill=cyan!10, inner sep=1pt, minimum height=17pt, minimum width=17pt]
        \tikzstyle{node2}=[rectangle,draw=black!55, fill=red!5,inner sep=1pt, minimum height=17pt, minimum width=30pt]
        \tikzstyle{node3}=[rectangle,draw=black!55, fill=cyan!10]
        \tikzstyle{dots}=[inner sep=0pt]
        \tikzstyle{model}=[rectangle, ,draw=black!55,  inner sep=1pt]
        
        \node[node] (t) at (0*\horisep,0*\vertsep) {$t$};
        \node[node] (x) at (0*\horisep,-1*\vertsep) {$x$};
        \node[node] (y) at (0*\horisep,-2*\vertsep) {$y$};
        \node[node] (s) at (0*\horisep,-3*\vertsep) {$s$};
        \node[node] (wx) at (0*\horisep,-4*\vertsep) {$w_x$};
        \node[node] (wy) at (0*\horisep,-5*\vertsep) {$w_y$};
        \node[rectangle, draw=black!55, fill=green!10, minimum height=5.5*\vertsep, minimum width=1cm] (nn) at (1*\horisep,-2.5*\vertsep) {FNN};
        \draw (t) -- (t -| nn.west);
        \draw (x) -- (x -| nn.west);
        \draw (y) -- (y -| nn.west);
        \draw (s) -- (s -| nn.west);
        \draw (wx) -- (wx -| nn.west);
        \draw (wy) -- (wy -| nn.west);
        
        \node[node1] (u) at (2*\horisep,-2.5*\vertsep) {$\tilde{u}$};
        \draw (nn) -- (u);
        
        \node[node2] (dx) at (3*\horisep,-1.5*\vertsep) {$\partial \tilde{u} / \partial x$};
        \node[node2] (dy) at (3*\horisep,-2.5*\vertsep) {$\partial \tilde{u} / \partial y$};
        \node[node2] (dt) at (3*\horisep,-3.5*\vertsep) {$\partial \tilde{u} / \partial t$};
        \draw (u) -- (dx);
        \draw (u) -- (dy);
        \draw (u) -- (dt);
        
        \node[node3] (pde) at (4.5*\horisep,-2.5*\vertsep) {$\frac{\partial \tilde{u}}{\partial t} + \vec{C} \cdot \left[ \frac{\partial \tilde{u}}{\partial x}, \frac{\partial \tilde{u}}{\partial y}  \right]^\top$};
        \draw (dx) -- (pde);
        \draw (dy) -- (pde);
        \draw (dt) -- (pde);

        \node[rectangle, dashed, draw=black!30, minimum height=6.5*\vertsep, minimum width=2.8*\horisep] () at (1*\horisep,-2.5*\vertsep) {};
        \node[rectangle, dashed, draw=black!30, minimum height=3.8*\vertsep, minimum width=3.7*\horisep] () at (3.5*\horisep,-2.5*\vertsep) {};
        \node[rectangle, dashed, draw=black!30, minimum height=7.4*\vertsep, minimum width=5.9*\horisep] () at (2.5*\horisep-1pt,-2.2*\vertsep) {};
        \node[] () at (1*\horisep,1*\vertsep) {Neural network};
        \node[] () at (3.5*\horisep,-0.35*\vertsep) {PDE};
        \node[] () at (2.5*\horisep-1pt,1.8*\vertsep) {PINN};
    \end{footnotesize}
\end{tikzpicture}
    \caption{Architecture of the proposed PINN. The FNN is a fully-connected feed-forward neural network that takes six inputs $t, x, y, s, w_x, w_y$ and predicts the level-set function $\tilde{u}$. The PDE is then constructed from partial derivatives that are calculated using automatic differentiation through the FNN. The PINN comprises both the neural network and the PDE.}
    \label{fig:pinn}
\end{figure}

The overall model structure of the PINN is illustrated in Figure \ref{fig:pinn}. The figure outlines the flow of information from the inputs of the neural network, to the prediction of the level-set function, to the formation of the PDE.

\subsubsection{PINN Likelihood}
\label{sec:pinnLikelihood}

The PINN parameters $\thetabf$ are fitted through gradient-based optimisation. For this, a set of data $\mathcal{D}$ and a cost function are required. In this study, the dataset comprises four subsets of data $\mathcal{D} = \{\mathcal{D}_i, \mathcal{D}_p, \mathcal{D}_f, \mathcal{D}_o \}$, where
\begin{align}
    \mathcal{D}_i &= \{ t_i^{(k)}, x_i^{(k)}, y_i^{(k)}, s_i^{(k)}, w_{x, i}^{(k)}, w_{y, i}^{(k)}, u_i^{(k)} \}_{k=1}^{N_i}, \\
    \mathcal{D}_p &= \{ t_p^{(k)}, x_p^{(k)}, y_p^{(k)}, s_p^{(k)}, w_{x,p}^{(k)}, w_{y,p}^{(k)} \}_{k=1}^{N_p}, \\
    \mathcal{D}_f &= \{ t_f^{(k)}, x_f^{(k)}, y_f^{(k)}, s_f^{(k)}, w_{x,f}^{(k)}, w_{y,f}^{(k)}, \hat{u}_f^{(k)} \}_{k=1}^{N_f}, ~ \text{and} \\
    \mathcal{D}_o &= \{ t_o^{(k)}, x_o^{(k)}, y_o^{(k)}, s_o^{(k)}, w_{x,o}^{(k)}, w_{y,o}^{(k)} \}_{k=1}^{N_o},
\end{align}
where $k$ denotes the $k^{\text{th}}$ sample in the dataset. The initial dataset $\mathcal{D}_i$ provides a set of initial (boundary) conditions, the physics dataset $\mathcal{D}_p$ comprises a set of collocation points\footnote{The term ``collocation points'' is used in PINN literature (e.g., \cite{raissi2019physics}) to describe samples from the spatio-temporal domain for the physics likelihood described in section \ref{sec:physicsLikelihood}.} \cite{raissi2019physics} over the spatio-temporal domain which are used to fit the model to the PDE (see Section \ref{sec:physicsLikelihood}), the forecast dataset $\mathcal{D}_f$ comprises a dataset with Euler predictions of the level-set function and is used for our proposed forecast likelihood, and the observation dataset $\mathcal{D}_o$ comprises a set of data used for data assimilation. The explanations of $t^{(k)}$, $x^{(k)}$, $y^{(k)}$, $u^{(k)}$, $\hat{u}^{(k)}$ for each subset are provided in subsequent sections. In all data subsets, $s^{(k)}$, $w_{x}^{(k)}$, and $w_{y}^{(k)}$ relate to external conditions at $(t^{(k)}, x^{(k)}, y^{(k)})$ and are provided by external sources, such as data from an anemometer for wind speed and direction. Finally, we denote the output of the neural network and the PINN for the $k^\text{th}$ input data sample $[t^{(k)}, x^{(k)}, y^{(k)}, s^{(k)}, w_{x}^{(k)}, w_{y}^{(k)}]^T$ by $\tilde{u}^{(k)}$ and $f^{(k)}$ respectively.

In the literature, the cost function used for optimising the PINN parameters $\thetabf$ is typically formed as a weighted sum of Mean Squared Errors (MSEs) as follows \cite{raissi2019physics, Karniadakis2021Physics}:
\begin{align}
    \label{eq:mseLoss}
    \mathcal{J}_{\text{cost}} = 
    \alpha_i \text{MSE}(\tilde{u}, \mathcal{D}_i) + \alpha_p \text{MSE}(f, \mathcal{D}_p) + \alpha_f \text{MSE}(\tilde{u}, \mathcal{D}_f) + \alpha_o \text{MSE}(\tilde{u}, \mathcal{D}_o),
\end{align}
where $\alpha_i$, $\alpha_p$, $\alpha_f$, and $\alpha_o$ are weighting factors that are treated as hyper-parameters in the optimisation process (and the MSEs relating to $\mathcal{D}_f$ and $\mathcal{D}_o$ are the novel additions proposed in this study). However, in the context of the B-PINN, we require a likelihood function rather than a MSE cost function. We thus model data by considering that the physics of the fire-front evolution defines mean values and that deviations between data and these means are i.i.d. Gaussian random variates. The PINN, which is parametrised by $\thetabf$, is optimised by maximising the likelihood given by (also see \cite{yang2021bpinns})
\begin{align}
    \label{eq:ltot}
    p(\mathcal{D} | \thetabf) = p(\mathcal{D}_i | \thetabf) p(\mathcal{D}_p | \thetabf) p(\mathcal{D}_f | \thetabf) p(\mathcal{D}_o | \thetabf).
\end{align}

The likelihoods $p(\mathcal{D}_i | \thetabf)$ and $p(\mathcal{D}_p | \thetabf)$ are referred to as the initial likelihood and the physics likelihood respectively \cite{raissi2019physics}. The forecast likelihood $p(\mathcal{D}_f | \thetabf)$ and the data observation likelihood $p(\mathcal{D}_o | \thetabf)$ are novel additions to the PINN optimisation objective as proposed in this study and are described in subsequent sections.

Given the general relationship between the MSE and the Gaussian log-likelihood in optimisation settings \cite{goodfellow2016deep}, $\ln p(\mathcal{D} | \thetabf)$ can be viewed as being in the form of the MSE cost function in (\ref{eq:mseLoss}) (plus a constant), where the variances in the likelihood function are related to the weighting factors in the cost function. However, the variances may be unknown; particularly for the synthetic data in $\mathcal{D}_p$ and $\mathcal{D}_f$ described in sections \ref{sec:physicsLikelihood} and \ref{sec:causalLoss} respectively. We thus follow the approach in (\ref{eq:mseLoss}) and treat unknown variances as hyper-parameters in the optimisation process such that they operate as weighting factors of the components in the log-likelihood function. As hyper-parameters, the unknown variances can be selected by either grid searches over some predefined range (as we do in this study), by cross-validation (which is rarely performed when model processing time is significant), or by multi-objective optimisation \cite{rohrhofer2021pareto}. Note that, when the variances are treated as hyper-parameters, the approach is not fully Bayesian and is more closely related to empirical Bayesian approaches where the B-PINN can be viewed as an empirical hierarchical model.

\subsubsection{Initial Likelihood}
\label{sec:initialLikelihood}

The dataset $\mathcal{D}_i$ comprises initial conditions (boundary conditions typically do not exist in the wildfire context) and it guides the neural network to learn the initial shape of $u(t,x,y)$. The dataset $\mathcal{D}_i$ is constructed by generating a set of $N_i$ samples of $u$ from a signed-distance function (such as (\ref{eq:cone}) or (\ref{eq:ellipticalCone}) discussed in section \ref{sec:signedDistanceFunction}) evaluated at a set of $N_i$ locations in the spatial domain. These locations could be specified by a discrete grid over the spatial domain or by randomly sampling over the spatial domain. If we assume that the outputs of the neural network are Gaussian distributed with a standard deviation of $\sigma_i$, the initial likelihood for $\mathcal{D}_i$ is given by
\begin{align}
    \label{eq:L_initial}
    p(\mathcal{D}_i | \thetabf) = \prod_{k=1}^{N_i} \frac{1}{\sqrt{2 \pi \sigma_i^2}} \exp \left( - \frac{ \left( \tilde{u}^{(k)} - u_i^{(k)} \right)^2 }{ 2 \sigma_i^2} \right).
\end{align}
The standard deviation $\sigma_i$ relates to the error in the initial conditions data. If the initial conditions are based on an observed fire-front, $\sigma_i$ relates to the noise in these observations.

\subsubsection{Physics Likelihood}
\label{sec:physicsLikelihood}

The physics likelihood $p(\mathcal{D}_p | \thetabf)$ has the purpose of minimising $f(t,x,y,s,w_x,w_y; \thetabf)$ in (\ref{eq:pinn}). If $f(t,x,y,s,w_x,w_y; \thetabf)$ is minimised to zero, the neural network has solved the level-set equation in (\ref{eq:levelSetEquation}) and it is constrained to follow the spatio-temporal dynamics (or so-called ``physics'') of the PDE. The dataset $\mathcal{D}_p$ comprises a set of collocation points \cite{raissi2019physics} over the spatio-temporal domain considered. Each collocation point $(t_p^{(k)}, x_p^{(k)}, y_p^{(k)})$ may be sampled randomly within the domain or sampled from a predefined spatio-temporal grid. The collocation point (along with the $s_p^{(k)}$, $w_{x,p}^{(k)}$, and $w_{y,p}^{(k)}$ values associated with this point) are passed through the PINN to produce $f^{(k)}$. As the solution to the PDE at this collocation point is $f(t_p^{(k)},x_p^{(k)},y_p^{(k)},s^{(k)},w_{x,p}^{(k)},w_{y,p}^{(k)}; \thetabf)=0$, we assume a zero-mean Gaussian with a standard deviation of $\sigma_p$ and the physics likelihood is given by
\begin{align}
    \label{eq:L_physics}
    p(\mathcal{D}_p | \thetabf) = \prod_{k=1}^{N_p} \frac{1}{\sqrt{2 \pi \sigma_p^2}} \exp \left( - \frac{ \left( f^{(k)} \right)^2 }{ 2 \sigma_p^2} \right).
\end{align}
Note that targets (or response variables\footnote{The target or label in machine learning is the same as a response variable in statistics and refers to the expected output of a model for a given input. In supervised machine learning, they are provided in a dataset along with their associated inputs (sometimes called covariates in statistics).}) of the level-set function $u$ are not required in $\mathcal{D}_p$ as they were in $\mathcal{D}_i$. The PDE functions as a target and the PINN is minimised such that the neural network models the solution to the PDE. This can be viewed as a form of semi-supervised training.

As $\mathcal{D}_p$ is a simulated dataset, $\sigma_p$ is unknown and is treated as a hyper-parameter. To represent a perfect solution to the PDE, $\sigma_p$ should ideally be zero to force the Gaussian into a Dirac distribution at the origin. However, the PINN is an approximation and is not guaranteed to find a perfect solution. Furthermore, an arbitrarily small value can cause the optimiser to focus on maximising $p(\mathcal{D}_p | \thetabf)$ and ignore the other likelihoods in $p(\mathcal{D} | \thetabf)$. The value of this hyper-parameter is thus usually set larger than $\sigma_i$ to ensure that the optimiser sufficiently fits to the initial condition data $\mathcal{D}_i$. That is, the value of $\mathcal{D}_p$ is tuned to ensure that the PINN follows the dynamics of the PDE and neural network fits well to the initial conditions.

\subsubsection{Forecast Likelihood and Temporal Continuity}
\label{sec:causalLoss}

Whilst the initial likelihood and the physics likelihood are important attributes that we wish to maximise, these are often not enough to ensure that a PINN is trained to produce physically realistic dynamics for fire-front evolution through time.  For example, it is possible for the a PINN optimiser to find a degenerate solution to the PDE that simply sets all derivatives to zero. As we show in our results, the PINN can yield such degenerate solutions when there are abrupt changes in $\vec{C}$. 

To address this problem, we propose an approach that encourages the model to maintain temporal continuity of the level-set function. This is achieved by comparing the neural network prediction of level-set function at a current time $t$ with an independent forecast of the current time based on the neural network prediction of the level-set function at a previous time $t - \Delta t$. This introduces a temporal conditioning over time into the neural network which encourages temporal continuity. The forecast is calculated based on the reformulation of (\ref{eq:levelSetEquation}) into the integral form of
\begin{align}
    u(t,x,y) = u(t - \Delta t,x,y) + \Delta t  \int_0^1 \vec{C} \cdot \nabla u(t,x,y) dt.
\end{align}
The Euler approximation to this integral is
\begin{align}
    \label{eq:prediction_}
    u(t,x,y) \approx u(t - \Delta t,x,y) + \Delta t \vec{C} \cdot \nabla u(t - \Delta t,x,y).
\end{align}

A dataset $\mathcal{D}_f$ is created by generating a set of points over space and time, where time is discretised into steps of size $\Delta t$ and the dataset is ordered with increasing time. That is, for any time $t^{(k)}$ indexed by $k$ in $\mathcal{D}_f$, $t^{(k)} > t^{(k-1)}$ and $t^{(k)} - t^{(k-1)} = \Delta t$. Replacing $u(t^{(k-1)},x,y)$ with the neural network approximation $\tilde{u}^{(k-1)}$, we have that
\begin{align}
    \label{eq:prediction}
    \hat{u}_f^{(k)} = \tilde{u}^{(k-1)} + \Delta t \vec{C} \cdot \hat{\n} \nabla \tilde{u}^{(k-1)},
\end{align}
where $\hat{u}_f^{(k)} \approx u(t_f^{(k)},x_f^{(k)},y_f^{(k)})$ is the Euler forecast of $u$ at time $t_f^{(k)}$. The output of the neural network for sample $k$ (which is associated with time $t_f^{(k)}$) is compared with the forecast that is computed from the output of the neural network for sample $k-1$ (which is associated with time $t_f^{(k-1)} = t_f^{(k)} - \Delta t$). This comparison is performed with forecast likelihood function given by
\begin{align}
    \label{eq:L_forecast}
    p(\mathcal{D}_f | \thetabf) 
    &= \prod_{k=2}^{N_p} \frac{1}{\sqrt{2 \pi \sigma_f^2}} \exp \left( - \frac{ \left( \tilde{u}^{(k)} - \hat{u}_f^{(k)} \right)^2 }{ 2 \sigma_f^2} \right), \\
    &= \prod_{k=2}^{N_p} \frac{1}{\sqrt{2 \pi \sigma_f^2}} \exp \left( - \frac{ \left( \tilde{u}^{(k)} - \left( \tilde{u}^{(k-1)} + \Delta t \vec{C} \cdot \hat{\n} \nabla \tilde{u}^{(k-1)} \right) \right)^2 }{ 2 \sigma_f^2} \right). \nonumber 
\end{align}
During training, this component of the overall likelihood function in (\ref{eq:ltot}) helps to penalise the types of degenerate solutions previously described. Consequently, training places importance on maintaining temporal continuity of the solution and producing physically realistic solutions. 

The spatial locations in the data samples are required to remain consistent over time to ensure that the $u^{(k)}$ and the forecast from $u^{(k-1)}$ are from the same point in space. To ensure this consistency, a spatio-temporal grid-based sampling scheme is used where the sample $k$ may comprise a set of sub-samples across the fixed spatial grid associated with time $t^{(k)}$. Furthermore, as the forecast is based on the output of the neural network for a previous sample $k-1$, the dataset is generated on-the-fly during training. Note that while the sampling scheme and the forecasts are discrete, the PINN solution is not and it remains in its functional form. 

The Euler approximation used for the forecast likelihood is a linear approximation and can be inaccurate with highly non-linear changes of $u$ in time. Though we did not encounter any problems associated with this, if necessary, the error can be reduced by reducing $\Delta t$ or by using more advanced integration approximations such as higher-order Runge-Kutta methods. Furthermore, the dataset $\mathcal{D}_f$ does not have to cover the entire domain of time considered as it may only be necessary to cover specific times or locations where the PINN tends to shift to a degenerate solution.

The standard deviation $\sigma_f$ represents the deviation between the neural network output for sample $k$ and the Euler forecast for sample $k$ and is unknown. As indicated in Section \ref{sec:pinnLikelihood}, we treat it as a hyper-parameter that controls the weighting of the forecast likelihood in the overall likelihood $p(\mathcal{D} | \thetabf)$. If $\sigma_f$ is set too small, the optimiser will force the neural network to follow the (crude) Euler approximation too closely. If set too large, the forecast likelihood will be dominated by the other likelihood components in (\ref{eq:ltot}) during optimisation and have no effect. The value of $\sigma_f$ is tuned to eliminate the degenerate solutions and the forecast likelihood is used in situations where degenerate solutions are encountered.

\subsubsection{Observation Likelihood and Data Assimilation}
\label{sec:dataAssimilationLoss}

In the wildfire context, data assimilation involves updating the level-set function based on observed locations of the fire-front. Such observations may be captured from fire fighters or remote sensing data such as satellite and drone images \cite{Schroeder2014New}. Suppose a dataset comprising $N_a$ point locations of the fire-front across space and time are provided. Each observation contains a location $(x,y)$ of a point on the fire-front, the time $t$ that the sample was observed, and the external conditions $\vec{C}$ at this location and time. As these observations lie on the fire-front, they lie on the zero-level-set and the level-set function should evaluate to zero at all of the points represented by the observations. We thus assume that the level-set function at the location of an observation is Gaussian-distributed with a zero mean and variance $\sigma_o^2$ such that the observation likelihood is given by
\begin{align}
    \label{eq:L_assimilation}
    p(\mathcal{D}_o | \thetabf) = \prod_{k=1}^{N_o} \frac{1}{\sqrt{2 \pi \sigma_o^2}} \exp \left( - \frac{ \tilde{u} \left(t^{(k)}_o, x^{(k)}_o, y^{(k)}_o, s_o^{(k)}, w_{x,o}^{(k)}, w_{y,o}^{(k)} \right)^2 }{ 2 \sigma_o^2} \right).
\end{align}
Maximising this likelihood component penalises solutions that fail to propagate the level-set function such that its zero-level-set passes through the observations.

The standard deviation $\sigma_f$ relates to the uncertainty in the observations in $\mathcal{D}_o$ and can be determined according to the noise in these observations. Note that the observations do not have to span the entire fire-front and they do not have to span across all time. Furthermore, the observations are not required to originate from the same source. For example, the observation at time $t^{(k)}$ is a single point determined by the location of a firefighter team and the observation at time $t^{(k+1)}$ is a set of points spanning the fire-front obtained from satellite imagery. However, in this case, $\sigma_f$ should be appropriately chosen for a data sample depending on the data source.

\subsubsection{Uncertainty Quantification and the Bayesian PINN}
\label{sec:bpinn}

To provide uncertainty quantification, we consider the Bayesian PINN (B-PINN) \cite{yang2021bpinns}. The premise of the B-PINN is to compute the posterior distribution over the parameters of the PINN, which is given by
\begin{align}
    p(\thetabf | \mathcal{D}) = \frac{p( \mathcal{D} | \thetabf) p(\thetabf)}{p(\mathcal{D})}.
\end{align}
As $p(\mathcal{D})$ is generally intractable in neural networks, we resort to finding an approximation to the posterior using variational inference. In particular, we follow the Bayes-by-backprop approach \cite{blundell2015weight}. Consider the variational posterior density function $q_{\phibf}(\thetabf)$, parametrised by $\phibf$, which provides an approximation to the posterior density $p(\thetabf | \mathcal{D})$. The variational inference approach optimises the parameters $\phibf$ to minimise the negative evidence lower bound (ELBO) given by
\begin{align}
    \label{eq:elbo}
    -\mathfrak{L}(q) = \mathbb{E}_{q_{\phibf}(\thetabf)}[ \ln q_{\phibf}(\thetabf) - \ln p(\mathcal{D}| \thetabf) -  \ln p(\thetabf)].
\end{align}
The mean-field approximation is assumed and the variational posterior and is given by
\begin{align}
    q_{\phibf}(\thetabf) = \prod_k \mathcal{N}(\thetabf_k | \mu_k, \sigma_k),
\end{align}
where $\mathcal{N}(\cdot)$ denotes a Gaussian probability density function and $\phi_k = (\mu_k, \sigma_k)$. The likelihood $p(\mathcal{D}| \thetabf)$ is defined in (\ref{eq:ltot}) and the prior $p(\thetabf)$ is assumed to take a spike-and-slab form, which is a Stochastic Search Variable Selection (SSVS) prior given by \cite{blundell2015weight}
\begin{align}
    \ln p(\thetabf) = \prod_k \frac{1}{2} \mathcal{N}(\theta_k | 0, \sigma_1) + \frac{1}{2} \mathcal{N} (\theta_k | 0, \sigma_2),
\end{align}
where the recommended values of $\sigma_1 = 1$ and $\sigma_2 = \exp(-6)$ are used in this study \cite{blundell2015weight}. 

It is important to note that Bayesian inference in neural networks is non-trivial due to the non-linearity and black-box nature of these models. Specifying a meaningful prior is challenging and the Gaussian prior is the standard choice given its mathematical properties \cite{goan2020bayesian}. Similarly, the mean-field assumption is a standard choice given its factorisation properties which reduce computational complexity and only increase the overall number of parameters in the PINN by a factor of two (each parameter is represented by a mean and variance). However, care should be taken as the mean-field assumption can tend to underestimate the true variance due to the independence assumptions across parameters \cite{blei2017variational}.

\subsubsection{Model Configurations and Nomenclature}

For clarity in subsequent sections, we assign names to the particular PINN configurations used in this study. These configurations are constructed based on variations in the likelihood function and the use of the Bayesian inference as follows: 

\noindent PINN-e is the elementary PINN defined in the literature, and its likelihood is given by
\begin{align}
    p(\mathcal{D} | \thetabf) = p(\mathcal{D}_i | \thetabf) p(\mathcal{D}_p | \thetabf).
\end{align}
PINN-f includes the forecast likelihood in (\ref{eq:L_forecast}) such that
\begin{align}
    p(\mathcal{D} | \thetabf) = p(\mathcal{D}_i | \thetabf) p(\mathcal{D}_p | \thetabf) p(\mathcal{D}_f | \thetabf).
\end{align}
PINN-a includes the forecast likelihood in (\ref{eq:L_forecast}) and the observation likelihood in (\ref{eq:L_assimilation}) such that
\begin{align}
    p(\mathcal{D} | \thetabf) = p(\mathcal{D}_i | \thetabf) p(\mathcal{D}_p | \thetabf) p(\mathcal{D}_f | \thetabf) p(\mathcal{D}_o | \thetabf).
\end{align}
The B-PINN-a includes all likelihoods and is the Bayesian counterpart of PINN-a. The B-PINN-f includes all likelihoods except the observation likelihood and is the counterpart of the PINN-f. The B-PINN models are optimised using variational inference and the remaining models are optimised using maximum likelihood optimisation.

\subsubsection{Optimisation and Application}
\label{sec:optimisation}

The standard gradient-descent based optimisation approach used for neural networks (commonly referred to as backpropagation) is used to train the PINN. The approach involves passing inputs through the neural network and comparing the outputs with targets via a cost function. The gradient of the cost with respect to the model parameters (weights and biases) is calculated layer-by-layer starting from the output layer and is used to update the parameters. This process is performed over a dataset comprising input-target pairs and repeated over several iterations called epochs. In this study, the cost function is either configured for maximum likelihood or for Bayesian inference.

For maximum likelihood optimisation, the PINN is trained to maximize the log-likelihood $\ln p(\thetabf | \mathcal{D})$ and is used for the PINN-e, PINN-f, and PINN-a models. The fitted neural network can be seen as a functional form of a surface over the domain of $(t,x,y)$ that represents the solution to the PDE. This solution can be evaluated by providing a point $(t,x,y)$ to the neural network to return approximation of $u(t,x,y)$. To illustrate the training procedure, a gradient ascent algorithm for the PINN-a is provided in Algorithm \ref{alg:pinn}. For the interested reader, illustrative examples where the PINN is applied to simple problems can be found in \cite{Karniadakis2021Physics}, \cite{Kollmannsberger2021Physics}, and \cite{raissi2017physics}.

For Bayesian inference in the B-PINN, the parameters $\phibf$ are optimised using backpropagation \cite{kingma2014autoencoding}. Compared to Algorithm \ref{alg:pinn}, the parameters $\phibf$ are updated according to the ELBO in (\ref{eq:elbo}) rather than updating $\thetabf$ according to the likelihood in (\ref{eq:ltot}). To make a prediction, a set of $N_{\text{MC}}$ Monte Carlo (MC) samples of the neural network parameters $\thetabf$ can be drawn from the variational posterior distribution. Each sample of $\thetabf$ can be used to produce a new simulation, generating a set of $N_{\text{MC}}$ MC simulations. The distribution over these simulations provides a MC estimate of the posterior predictive distribution. Statistics such as mean, standard deviation, and confidence intervals can be computed from these MC simulations to quantify the uncertainty of the PINN predictions.

\begin{algorithm}[!t]
        \begin{algorithmic}[1]
            \Require the PINN model (\ref{eq:pinn}) with its neural network (\ref{eq:mlp}) and parameters $\thetabf$; datasets $\mathcal{D}_i$, $\mathcal{D}_p$, $\mathcal{D}_f$, and $\mathcal{D}_o$; learning rate $\eta$; and number of epochs $M$.
            \State Randomly initialise the parameters $\thetabf$.
            \For{each epoch in $M$}
                \State Pass $\mathcal{D}_i$ through the neural network and compute $\ln p(\mathcal{D}_i | \thetabf)$ in (\ref{eq:L_initial}).
                \State Pass $\mathcal{D}_p$ through the PINN and compute $\ln p(\mathcal{D}_p | \thetabf)$ in (\ref{eq:L_physics}).
                \State Pass $\mathcal{D}_f$ through the neural network and store $\tilde{u}^{(k)}$ for each sample $k$.
                \State Initialise $\ln p(\mathcal{D}_f | \thetabf) = 0$ .
                \For{$k = 2$ to $N_f$}
                    \State Compute the forecast $\hat{u}^{(k)}$ using (\ref{eq:prediction}) with $\tilde{u}^{(k-1)}$.
                    \State Compute and update $\ln p(\mathcal{D}_f | \thetabf)$ for sample $k$ using (\ref{eq:L_forecast}) with $\hat{u}^{(k)}$ and $\tilde{u}^{(k)}$.
                \EndFor
                \State Pass $\mathcal{D}_o$ through the neural network and compute $\ln p(\mathcal{D}_o | \thetabf)$ in (\ref{eq:L_assimilation}).
                \State Compute the negative log-likelihood $-\ln p(\mathcal{D} | \thetabf)$ from (\ref{eq:ltot}).
                \State Compute the gradients $-\frac{\partial \ln p(\mathcal{D} | \thetabf)}{\partial \thetabf}$ using automatic differentiation.
                \State Update the model parameters using backpropagation according to $\thetabf \leftarrow \thetabf + \eta \frac{\partial \ln p(\mathcal{D} | \thetabf)}{\partial \thetabf}$.
            \EndFor
        \end{algorithmic}
    \caption{Maximum likelihood training algorithm for the PINN-a with gradient descent. Notes: (i) it is assumed that $\mathcal{D}_f$ contains $N_f$ samples ordered in time, where each sample contains a set of sub-samples across a fixed spatial grid and (ii) the more advanced ADAM optimisation algorithm \cite{kingma2014adam} is used in experiments rather than gradient descent.}
    \label{alg:pinn}
\end{algorithm}
\section{Results}

\subsection{Synthetic Dataset and Results}

A synthetic dataset is generated to demonstrate how PINN-e can fail to maintain temporal continuity and how the inclusion of the forecast likelihood $p(\mathcal{D}_f| \thetabf)$ can address this problem. The dataset follows a circular fire-front that propagates around an obstacle and undergoes an extreme change in wind direction. We show that the PINN-e loses track of the fire-front and violates temporal continuity as the wind changes direction, whereas the PINN-f is able to maintain the temporal continuity throughout the simulation. Additionally we adjust the synthetic dataset into a more complex form to demonstrate that the B-PINN-f is able to produce complex fire-front shapes.

\subsubsection{Synthetic Dataset}

A synthetic dataset is generated over a two-dimensional space $x\in[0,1]$, $y\in[0,1]$ and over the time period $t \in [0,1]$. The vector field $\vec{C}$ is varied in both of its components $\vec{S}$ and $\vec{W}$. To emulate an extreme change in wind direction, the wind vector direction is initially in the northerly direction for 10\% of the time, and it changes to an easterly direction for the remaining time. That is, for all $(x,y)$, the wind vector is given by
\begin{align}
    \vec{W}(t,x,y) = 
    \begin{cases}
        [0.0 ~~~ 0.4]^\top, & t \leq 0.1, \\
        [0.4 ~~~ 0.0]^\top, & t > 0.1.
    \end{cases}
\end{align}
The fire-front speed is assumed to be a constant of $s=0.4$ throughout space and time. A square-shaped obstruction is placed in the line of the propagating fire-front by enforcing the following values for the vector field $\vec{C}$, for all $t$:
\begin{align}
    \vec{C}(t,x,y) = 
    \begin{cases}
        \mathbf{0}, &  (x,y) \in \mathcal{A},  \\
        \vec{C}(t,x,y), & \text{otherwise}.
    \end{cases}
\end{align}
Here $\mathcal{A} = [0, 0.2] \times [0.2, 0.8]$. This obstruction could represent a region that cannot be burned, such as a water body.

\subsubsection{Complex Shape Synthetic Dataset}

For completeness, the synthetic dataset is also adjusted to demonstrate complex fire-front shapes. For this, three obstacles are produced such that, for all $t$:
\begin{align}
    \vec{C}(t,x,y) = 
    \begin{cases}
        \mathbf{0}, &  (x,y) \in \mathcal{B} \cap \mathcal{C} \cap \mathcal{D}, \\
        \vec{C}(t,x,y), & \text{otherwise}.
    \end{cases}
\end{align}

\noindent Here $\mathcal{B} = [0, 0.2] \times [0.2, 0,8]$, $\mathcal{C} = [0.7, 0.8] \times [0.4, 0,5]$, and $\mathcal{D} = [0.7, 0.8] \times [0.6, 0,7]$.  The wind vector field is homogeneous across space, and follows a random walk in the northerly direction such that
\begin{align}
    \vec{W}^{(k+1)} = \vec{W}^{(k)} + \mathcal{N} \left( \mathbf{0}, \text{diag}\left([0.001^2, 0.005^2]^\top \right) \right),
\end{align}
where $\vec{W}^{(0)} = [10^{-6}, 0.1]^\top$. Finally, the speed of the fire-front is split over the $x-$domain such that, for all $t$:
\begin{align}
    \vec{S}(t,x,y) = 
    \begin{cases}
        0.25, & x<0.5, \\
        0.15, & x \geq 0.5.
    \end{cases}
\end{align}
Spatial non-homogeneity of the fire spread in response to differences in fuel load or topography is achieved by allowing  $\vec{S}(t,x,y)$ to vary across space.  In addition, basic temporal non-homogeneity in wind direction is enforced with the time varying $\vec{W}(t,x,y)$. This adjusted synthetic dataset is referred to as Synthetic2.

\subsubsection{Methodology}

A spatio-temporal grid is created over the range of the data such that the spatial step sizes are $\Delta x = \Delta y = 1/35$ and the temporal step size is $\Delta t = 1/48$. This spatio-temporal grid is used to sample collocation points for the PINNs and it is used as the level-set method grid. The initial fire perimeter is configured as a circle with a radius of $0.1$ using (\ref{eq:cone}) with $r = -0.1$.

The hyper-parameters for the PINN-e and PINN-f are provided in the first row of Table \ref{table:pinn_config}. Other than the inclusion of $\sigma_f$ in the PINN-f, the hyper-parameters are identical between the two models. The hyper-parameters for B-PINN-f, which is tested on Synthetic2 are provided in the second row of Table \ref{table:pinn_config}. Hyper-parameters are empirically selected using grid searches where the learning rate is searched over a logarithmic scale in the range $[10^{-6}, 10^{-1}]$ and the variances are searched over a logarithmic scale in the range $[(2\pi10^4)^{-1}, (2\pi)^{-1}]$. The ADAM algorithm \cite{kingma2014adam} (a gradient-descent based algorithm) is used to minimise the negative log-likelihood of the models. The level-set method is used without reinitialisation for the synthetic datasets.
\begin{table}[!t]
    \centering
    \caption{PINN and B-PINN Configurations for the various datasets. $N_t$, $N_x$, and $N_y$ define the sample grid size where $N_t$ is the number of time segments, $N_x$ is the number of segments in the $x-$direction, and $N_y$ is the number of segments in the $y-$direction. The learning rate is denoted by l.r. The layer dimensions (layer dims.) include the input, the two hidden, and the output layer dimensions.}
    \label{table:pinn_config}
    \begin{scriptsize}
        \begin{tabular}{ccccccccccc}
            \toprule
            Dataset 	& Layer dims. 		& $N_t$	& $N_x$	& $N_y$	& l.r. 		& Epochs 	& $\sigma_i^2$	& $\sigma_p^2$	& $\sigma_f^2$	& $\sigma_o^2$ \\
            \midrule
            Synthetic	& [6, 64, 64, 1] 	& 48 	& 35 	& 35 	& $1e-3$ 	& 6 000 	& $\frac{1}{2 \times 1000}$ 	& $\frac{1}{2}$		& $\frac{1}{2 \times 50}$ 	&  $\frac{1}{2 \times 1000}$ \\
            Synthetic2	& [6, 64, 64, 1] 	& 48 	& 35 	& 35 	& $1e-3$ 	& 16 000 	& $\frac{1}{2 \times 1000}$ 	& $\frac{1}{2}$		& $\frac{1}{2 \times 50}$ 	&  $\frac{1}{2 \times 1000}$ \\
            Fire S03	& [6, 128, 128, 1] 	& 68 	& 71 	& 71 	& $5e-4$ 	& 50 000 	& $\frac{1}{2 \times 1000}$		& $\frac{1}{2}$		& $\frac{1}{2 \times 50}$ 	& $\frac{1}{2 \times 1000}$ \\
            Fire E06	& [6, 128, 128, 1] 	& 69 	& 57 	& 57 	& $1e-4$ 	& 50 000 	& $\frac{1}{2 \times 1000}$		& $\frac{1}{2}$		& $\frac{1}{2 \times 50}$ 	& $\frac{1}{2 \times 1000}$ \\
            \bottomrule
        \end{tabular}
    \end{scriptsize}
\end{table}

The dataset is constructed to contain three subsets: the initial condition data $\mathcal{D}_i$ at time $t=0$; the collocation points $\mathcal{D}_p$, which range over $t\in[0,1]$; and the forecast dataset $\mathcal{D}_f$, which is generated from the same collocation points in $\mathcal{D}_p$ during training. For all subsets, a set of samples are drawn over the spatio-temporal grid for the neural network inputs $t$, $x$, $y$, $s$, $w_x$, and $w_y$. These six inputs are concatenated and the set of samples over the spatio-temporal grid are collapsed into a single dimension to form one large batch (that is, full batch processing is used). For example, given the grid dimensions provided in Table \ref{table:pinn_config}, the dimensions of the neural network input for the collocation data are $[N_t \times N_x \times N_y, 6] = [48 \times 35 \times 35, 6] = [55800, 6]$, where $N_t$, $N_x$, and $N_y$ are the grid sizes of the temporal and spatial dimensions.

We found that the neural network was slow to converge to a solution that accurately represented the discontinuity of signed-distance function (i.e., the point of the cone) and tended to smooth it out. This can be problematic in accurately representing the zero-level-set, especially when it is small in size. We thus increase the number of the initial condition samples in $\mathcal{D}_i$ within the region bounded by the zero-level-set at time $t=0$ such that an additional $35\cdot35=1225$ samples are located from within this region. This was not performed on the collocation samples in $\mathcal{D}_p$ and $\mathcal{D}_f$.

The results are evaluated through visual plots and the Jaccard index. The Jaccard index is an intersection-over-union measure between two predicted burned regions $A$ and $B$ given by
\begin{align}
    \label{eq:jaccard}
    J(A,B) = \frac{|A \cap B|}{|A \cup B|}.
\end{align}
The burned regions are defined as the region where the level-set function is less than zero (below the zero-level-set).

\subsubsection{Wind Direction Change Results}

\begin{figure}[!t]
    \centering
    \begin{subfigure}[t]{5.39in}
        \includegraphics[width=5.39in]{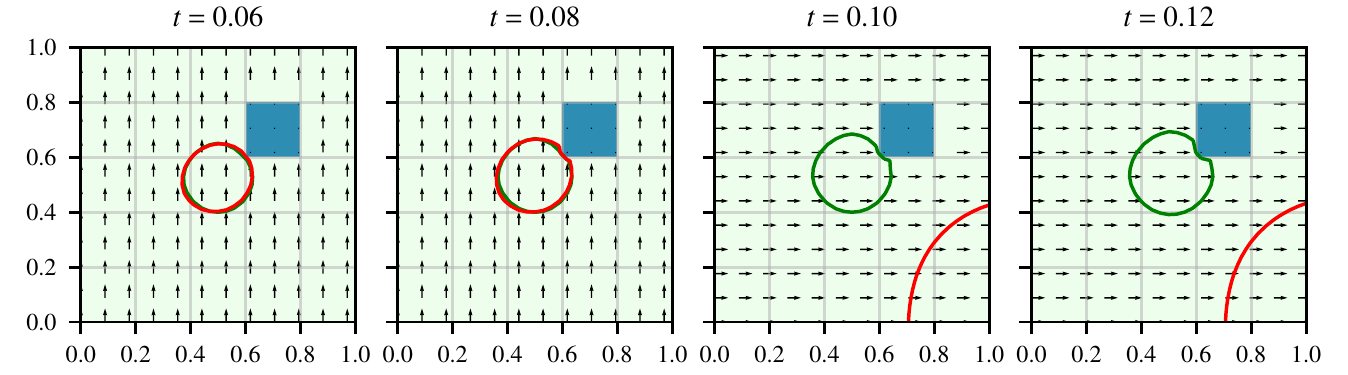}
        \caption{PINN-e predictions of the zero-level set. The red isochrone is the PINN-e prediction, the green isochrone is the level-set method prediction, and the blue square is the obstruction.}
        \label{fig:nopredcontours}
    \end{subfigure}
    \begin{subfigure}[t]{5.39in}
        \includegraphics[width=5.39in]{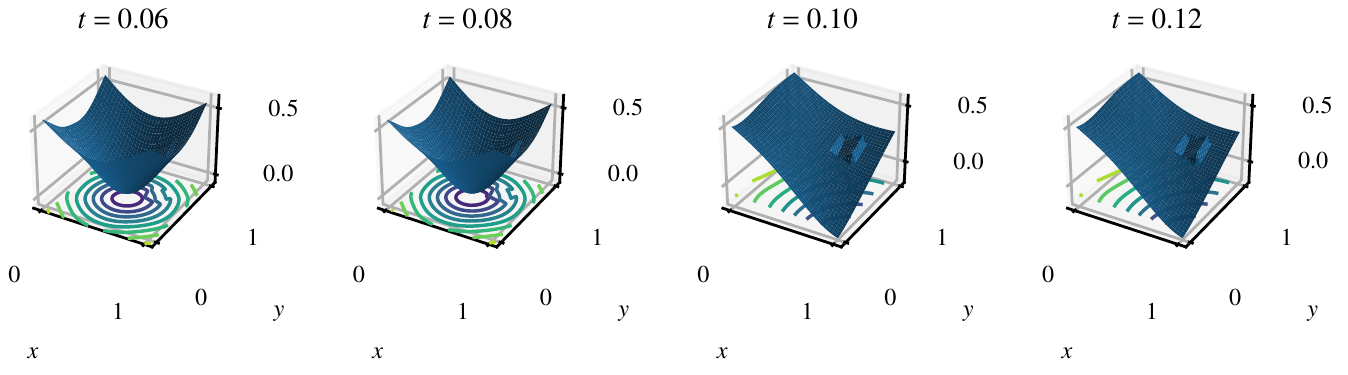}
        \caption{Surface plots of the PINN-e predictions of the level-set function.}
        \label{fig:noPredSurfaces}
    \end{subfigure}
    \caption{PINN-e (with no forecast likelihood) results for time steps around $t=0.1$ where the wind direction changes. The PINN-e takes advantage of the sudden change in wind direction at time t=0.1 to simplify the level-set function and its spacial derivatives. This change however is not physically justified and is not temporally continuous.}
\end{figure}

\begin{figure}[!t]
    \centering
    \includegraphics[width=3.5in]{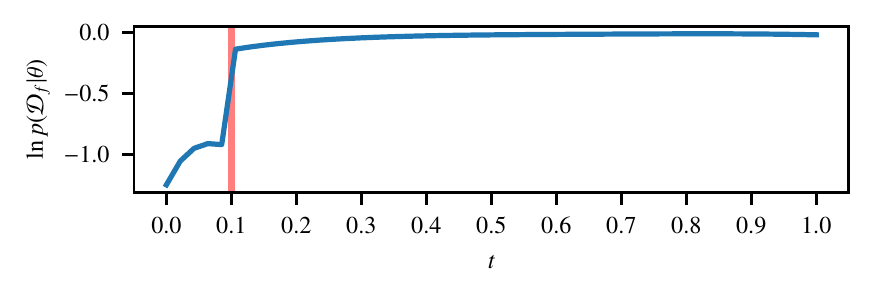}
    \caption{Physics log-likelihood over time. The red marker indicates the time that the wind changes direction. At this point the PINN optimiser conveniently changes the form of the level-set function, but in a way that is not physically realistic. See Figure \ref{fig:noPredSurfaces}.}
    \label{fig:windChangeLoss}
\end{figure}

The PINN-e results on the synthetic dataset are shown in Figure \ref{fig:nopredcontours}. The PINN-e solution initially follows the level-set method solution, but diverges as the wind changes direction at time $t=0.1$. As illustrated in Figure \ref{fig:noPredSurfaces}, the PINN-e level-set function changes from a cone to a simpler shape that is closer to a plane. This simplified shape has smaller spatial gradients which provide a more optimal solution to (\ref{eq:pinn}). This is illustrated in Figure \ref{fig:windChangeLoss} where the physics log-likelihood ($\ln p(\mathcal{D}_p | \thetabf)$) is plotted over time. As the wind changes direction at $t=0.1$, the log-likelihood abruptly increases to a value closer zero, where a value of zero would indicate an exact solution of the level-set equation. The PINN-e has thus taken advantage of the discontinuity in the wind direction to shift to a completely different solution, which from the perspective of the optimiser, is a more optimal solution. However this solution is not physically justified and the temporal continuity of the level-set function has not been maintained. In general, this temporal continuity problem is a well-known problem with PINN-e as discussed in sections \ref{sec:relatedWork} and \ref{sec:causalLoss}.

%
%

%
%

%
\begin{figure}[!t]
    \centering
    \includegraphics[width=5.39in]{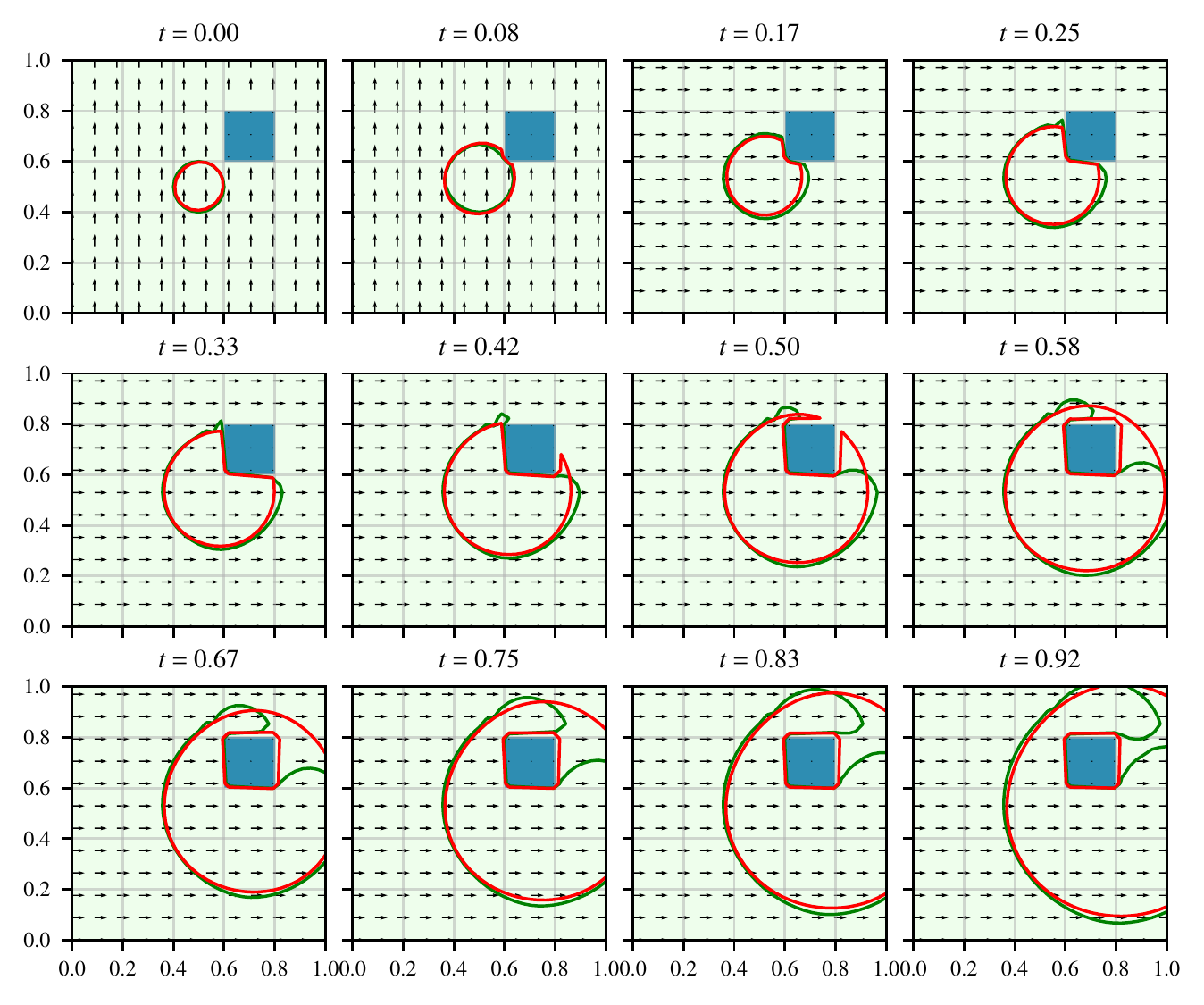}
    \caption{Synthetic dataset results with the PINN-f model (with the forecast likelihood). The red isochrone is the PINN-f prediction, the green isochrone is the level-set method prediction, and the blue square is the obstruction. Unlike the PINN-e, the PINN-f is able to maintain temporal continuity over the change in wind direction.}
    \label{fig:syntheticResults}
\end{figure}
%

%
\begin{figure}[t]
    \centering
    \includegraphics[width=3.5in]{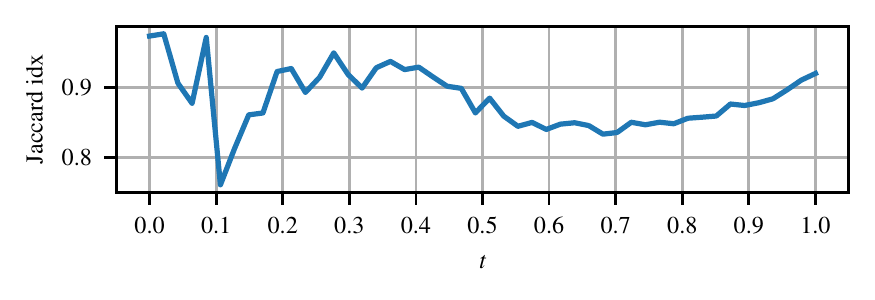}
    \caption{Jaccard index between the PINN and level-set method results over time for the synthetic dataset.}
    \label{fig:jaccardSynthetic}
\end{figure}

In contrast to the PINN-e, the results for the PINN-f and the level-set method are illustrated in Figure \ref{fig:syntheticResults}. With the inclusion of the forecast likelihood, the PINN-f adapts to the changes in the wind and maintains the general structure of the level-set function over time. By relating outputs of the neural network across time using the forecast likelihood, the PINN-f maintains temporal continuity in its solution. The Jaccard index between the PINN-f and level-set method are plotted over time in Figure \ref{fig:jaccardSynthetic}. As the wind changes direction, the Jaccard index drops to $76\%$, but remains above $80\%$ otherwise. This indicates a high level of similarity between the PINN-f and level-set method results.

A key difference between the PINN results and the level-set method results is in how they propagate the level-set function around the obstruction\footnote{Note that this problem is not only related to the PINN-f.}. In the level-set method, the propagation of a fire around the obstruction requires it to take a longer route than if it could pass through it.  Consequently, the fire takes a long time to reach the opposite side of the obstacle. Conversely, The PINN appears to pass the level-set function through the obstruction using the Euclidean route, maintaining its circular structure and reaching the opposite side of the obstacle faster. This highlights that the PINN can obtain physically implausible solutions. Unfortunately, inclusion of the forecast likelihood component in the overall likelihood does not correct this problem. However, the Jaccard index still remains well above $80\%$ as the level-set function passes the obstruction.

\subsubsection{Complex Fire-Front Shape Results}

\begin{figure}[!t]
    \centering
    \includegraphics[width=5.39in]{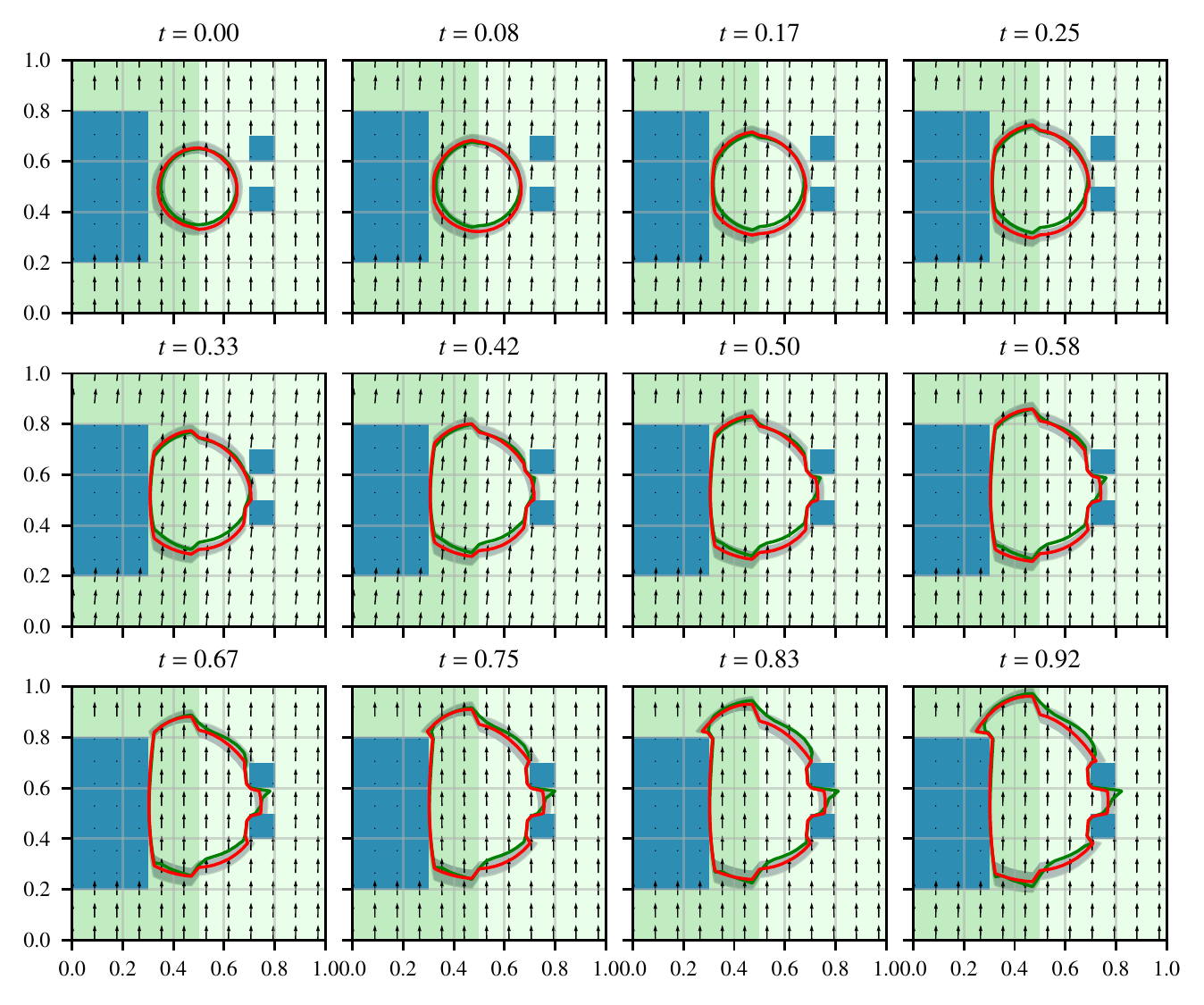}
    \caption{Synthetic dataset results with the B-PINN-f model (without assimilation) on the complex synthetic dataset Synthetic2. The red isochrone is the B-PINN-f prediction, the grey region indicates the 95\% posterior predictive interval, the green isochrone is the level-set method prediction, the quiver plot indicates the wind vector, and the blue regions are the obstructions. The left-hand side of the figure is a darker green to the right hand side to indicate that it has higher speed $s$.}
    \label{fig:syntheticResults_complex}
\end{figure}

The B-PINN-f model and the level-set method are applied to the adjusted synthetic dataset (Synthetic2) with the three obstructions and spatially varying speed. The results are illustrated in \figurename{} \ref{fig:syntheticResults_complex}. Like the level-set method, the B-PINN-f is able to follow the complex nature of the driving forces and the obstructions. In the left half of the region, the fire propagates faster owing to the higher speed $s$ in this region. Furthermore, with predominant wind in the northerly direction, the fire-front propagates faster in this direction. The result is that the B-PINN-f is able to represent complex shape changes and maintain results that are similar to what the level-set method produces over non-homogeneous conditions. Furthermore, the uncertainty quantification (represented by the 95\% posterior predictive interval) adapts to the environment where it is reduced near the obstacles where the fire-front is prevented from progressing. Evaluation of uncertainty quantification is conducted for the real-world datasets in Section \ref{sec:FireDatasetAndResults}.
%
%


\subsection{Application of PINNs to Australian Fires}
\label{sec:FireDatasetAndResults}

The PINN-f, the PINN-a, the B-PINN-a, and the level-set method are compared and contrasted on a dataset comprising two real-world fires. The aims are to (i) compare the three approaches, (ii) demonstrate data assimilation in the PINN, and (iii) demonstrate uncertainty quantification with the PINN.

\subsubsection{Fire Datasets}

\begin{figure}[!t]
    \centering
    \begin{subfigure}[t]{3.5in}
        \centering
        \includegraphics[width=3.5in]{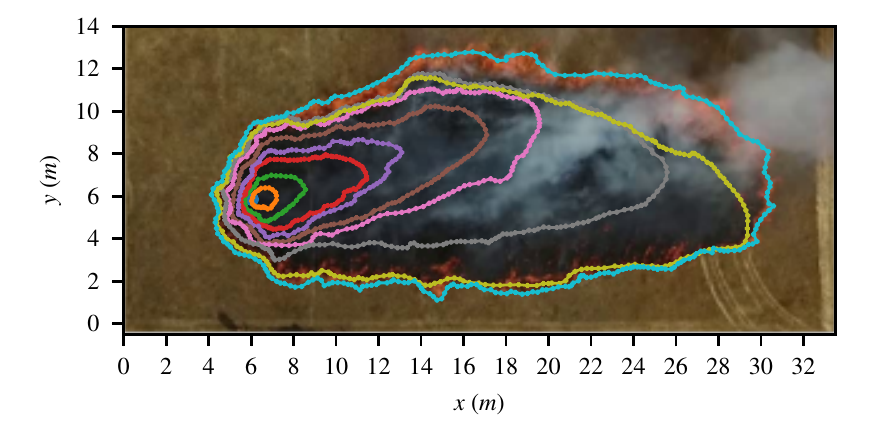}
        \caption{Fire S03.}
        \label{fig:braidwoodDataset_S03}
    \end{subfigure}
    \begin{subfigure}[t]{2.9in}
        \centering
        \includegraphics[width=2.9in]{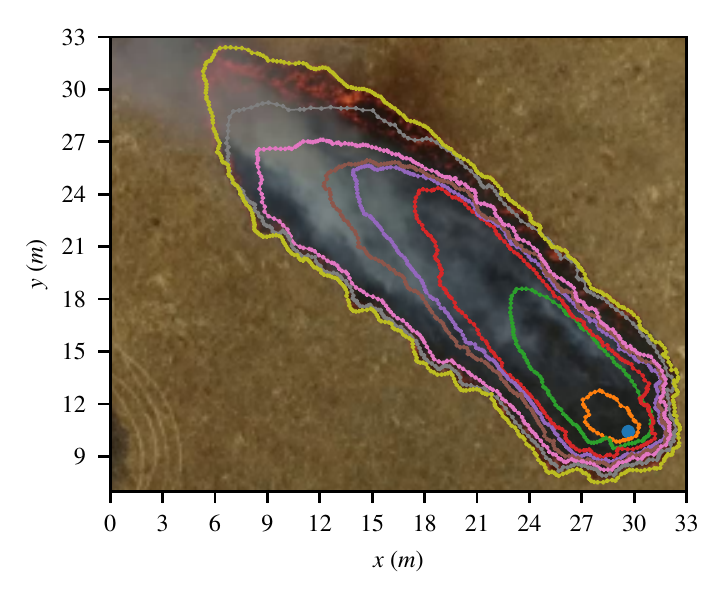}
        \caption{Fire E06.}
        \label{fig:braidwoodDataset_E06}
    \end{subfigure}
\caption{Depiction of two fires (S03 and E06) from the Braidwood dataset. Each image is the final fire state and is overlaid with fire-front isochrones showing the fire progression. The fire ignition is located at the blue isochrone (which appears as a blue dot), and the subsequent expanding isochrones depict the fire-front at 10 second intervals. In fire~S03 the wind speed and direction vary substantially over time. In fire~E06 there is a strong dominant wind blowing in the north-west direction. Axes are provided in meters ($m$).}
    \label{fig:braidwoodDataset}
\end{figure}

The Braidwood fire dataset \cite{sullivan2018study} is created from data collected from grassland fires which were filmed overhead near Braidwood, New South Wales, Australia. The two example fires, labelled S03 and E06, are illustrated in Figure \ref{fig:braidwoodDataset} and represent two types of fire regimes. Fire S03 houses complicated fire dynamics induced by variable wind speed and direction throughout the course of the fire. Fire E06 has strong dominant wind blowing in the north-west direction which results in an elongated elliptical fire-front. Although fire S03 is useful in demonstrating various aspects of the models, it initially accelerates and this acceleration is not modelled by the constant speed model in (\ref{eq:rothermal}). The result is a slight underestimation of the capability of the models to produce accurate results and the main purpose of fire E06 is thus to demonstrate the accuracy of the models.

The fire-fronts were manually mapped into isochrones at 10 second intervals based on the video frames from the overhead footage. Each fire-front isochrone comprises a set of samples of coordinates along the fire-front. Finally, wind speed and direction were measured with an anemometer located approximately 35 meters upwind from the ignition point and at a height of 2 meters.

\subsubsection{Methodology}

A spatio-temporal grid is initialised over the regions of the fires, with the fire ignition points located at the centres of the grids at time $t=0$. For fire S03, the spatial grid area is $71m \times 71m$ and for fire E06, the spatial grid area is $57m \times 57m$. For both fires, the spatial and temporal step sizes are $\Delta x = \Delta y \approx 1m$ and $\Delta t = 1s$ respectively. The spatio-temporal grid is scaled to a range of $[0,1]$ for both the spatial and temporal dimensions. The speed $s$ and wind $\vec{W}$ components are scaled according to the scaling of the spatio-temporal grid; except when applied as inputs to the neural network, where they are scaled to a range of $[0,1]$. The initial fire perimeter is set as an ellipse using the elliptical cone given by (\ref{eq:ellipticalCone}) with $a^2=5$, $b^2=1$, $k=-0.02$, and $\alpha=30^\circ$ for fire S03 and $a^2=0.5$, $b^2=7$, $k=-4.5$, and $\alpha=222^\circ$ for fire E06. The ellipse of S03 overestimates the initial fire size to account for the initial acceleration of the fire before it reaches a steady state. This results in poorer predictions in the early stages of the fire and the first few seconds are excluded in some of the results where noted.

The initial value dataset $\mathcal{D}_i$ is constructed according to the elliptical cones and the collocation dataset $\mathcal{D}_p$ is constructed on the spatio-temporal grid. The forecast dataset $\mathcal{D}_f$ is constructed from the collocation points in $\mathcal{D}_p$ during training. The observation dataset $\mathcal{D}_o$ is constructed based on the fire isochrones shown in Figure \ref{fig:braidwoodDataset}. As for the synthetic dataset, the region bounded by the zero-level-set is oversampled for the initial value dataset, with an additional $5041$ samples ($71 \times 71$) to ensure that the PINN models the discontinuity at the point of the elliptical cone.

The six neural network inputs $t$, $x$, $y$, $s$, $w_x$, and $w_y$ are concatenated and the set of samples over the spatio-temporal grid are collapsed into a single dimension. For example, given the grid size provided in Table \ref{table:pinn_config}, the dimensions of the neural network input for the collocation data for fire S03 are $[N_t \times N_x \times N_y, 6] = [68 \times 71 \times 71, 6] = [342788, 6]$.

The hyper-parameters for the PINN-f, the PINN-a and the B-PINN-a (which are identical to the PINN-a) are provided in the third and fourth rows of Table \ref{table:pinn_config} for datasets S03 and E06 respectively. These hyper-parameters were empirically selected using grid searches where the learning rate was searched over a logarithmic scale in the range $[10^{-6}, 10^{-1}]$ and the variances were searched over a logarithmic scale in the range $[(2\pi10^4)^{-1}, (2\pi)^{-1}]$. To provide some form of cross validation, we use the same likelihood variance parameters for both datasets. The ADAM algorithm \cite{kingma2014adam} is used to optimise the negative log-likelihood of the model and to optimise the variational parameters of the B-PINN-a. The level-set method reinitialisation is performed every $T_r = 10$ time-steps with one integration iteration.

The approaches are evaluated using plots and the Jaccard index given by (\ref{eq:jaccard}). The B-PINN-a uncertainty quantification is evaluated according to the ground-truth fire isochrones. At each 10 second interval, the 95\% posterior predictive interval is calculated from a set of 100 MC simulations drawn from the posterior predictive distribution of the B-PINN-a. The observed coverage is computed as the average number of ground-truth samples that lie within the 95\% confidence interval. We consider coverages that are close to the nominal 95\% value as showing a realistic representation of predictive uncertainty.

\subsubsection{Fire S03: Complicated Fire Dynamics}

\begin{figure}[!p]
    \centering
    \begin{subfigure}[t]{5.39in}
        \includegraphics[width=5.39in]{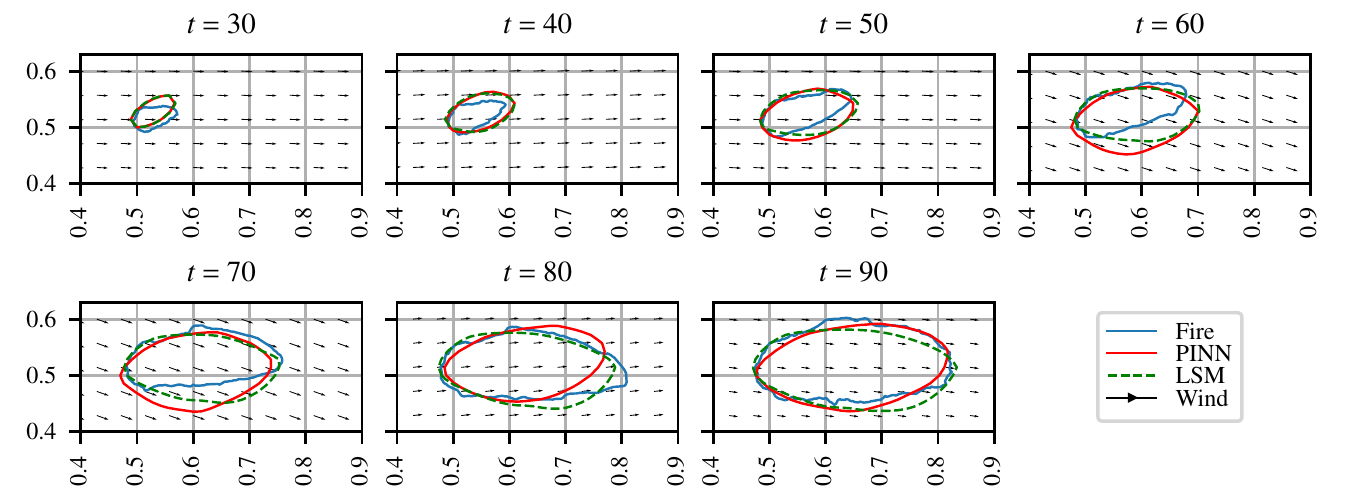}
        \caption{PINN-f versus the level-set method.}
        \label{fig:fireResults_noAssimilation}
    \end{subfigure}
    \begin{subfigure}[t]{5.39in}
        \includegraphics[width=5.39in]{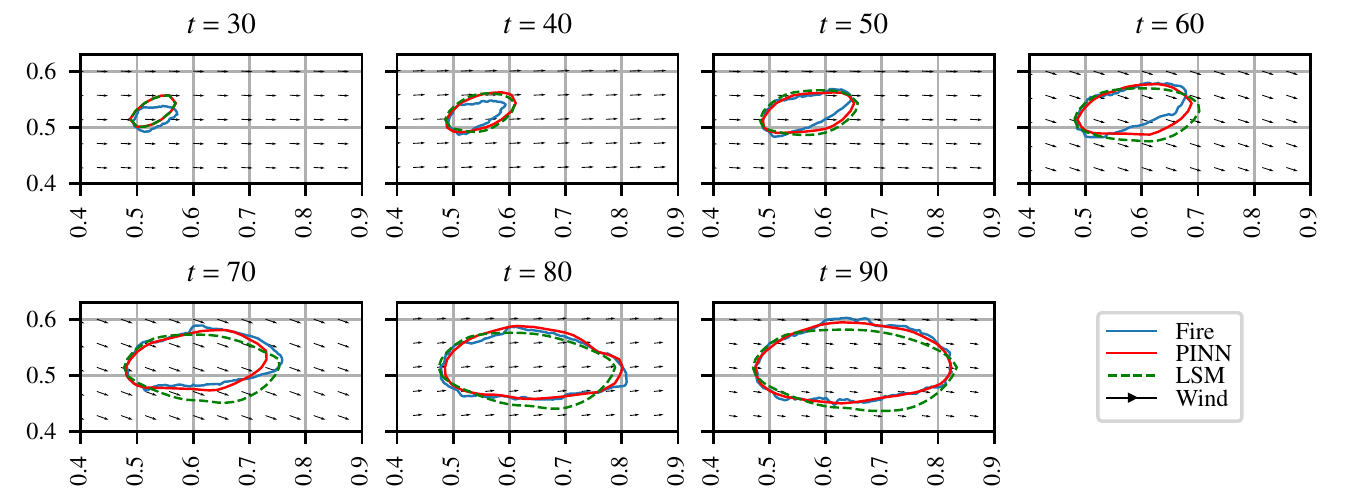}
        \caption{PINN-a versus the level-set method.}
        \label{fig:fireResults_assimilation}
    \end{subfigure}
    \begin{subfigure}[t]{5.39in}
        \includegraphics[width=5.39in]{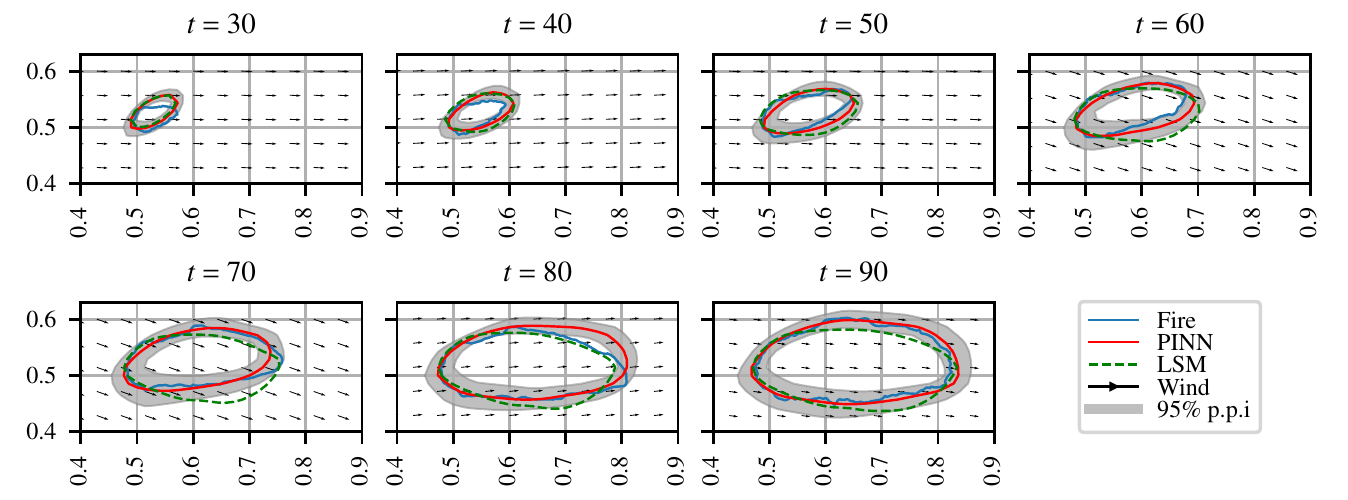}
        \caption{B-PINN-a versus the level-set method.}
        \label{fig:fireResults_assimilation_bpinn}
    \end{subfigure}
\caption{PINN-f, PINN-a, B-PINN-a, and level-set method results on the S03 fire dataset. LSM denotes the level-set method and p.p.i. denotes posterior predictive interval.}
    \label{fig:fireResults}
\end{figure}

The fire-front predictions for PINN-f and the level-set method are plotted in Figure \ref{fig:fireResults_noAssimilation}. While both approaches produce similar results,  neither is able to exactly reproduce the evolution of the actual fire-front. This is expected as the methods assume a homogeneous fuel type, fuel load, wind speed, and wind direction, which vary in reality. Furthermore, the fire initially accelerates and this acceleration is not accounted for in the constant speed model of (\ref{eq:rothermal}).

\begin{figure}[!t]
    \centering
    \begin{subfigure}[t]{3.2in}
        \includegraphics[width=3.2in]{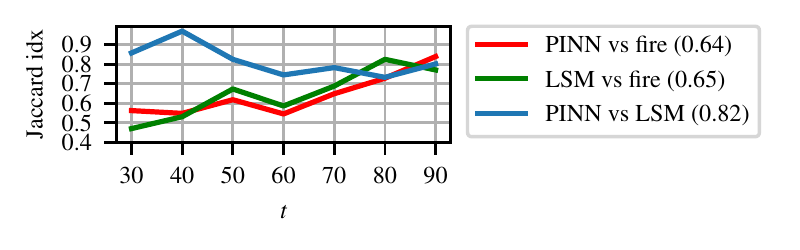}
        \caption{Fire S03: PINN-f versus the level-set method.}
        \label{fig:jaccardFire_noAssimilation}
    \end{subfigure}
    \begin{subfigure}[t]{3.2in}
        \includegraphics[width=3.2in]{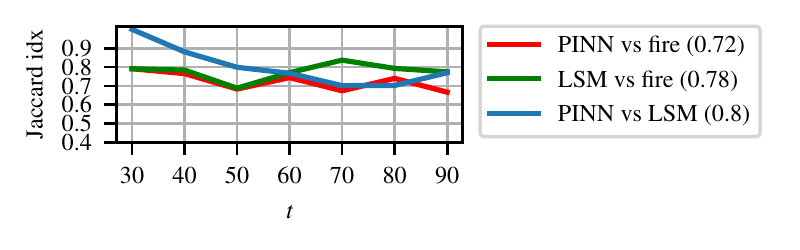}
        \caption{Fire E06: PINN-f versus the level-set method.}
        \label{fig:jaccardFire_noAssimilation_E06_Braidwood}
    \end{subfigure}
    \begin{subfigure}[t]{3.2in}
        \includegraphics[width=3.2in]{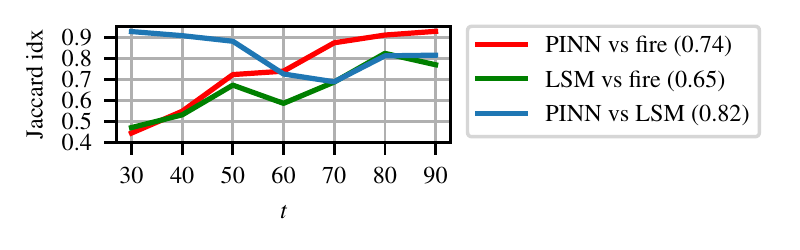}
        \caption{Fire S03: PINN-a versus the level-set method.}
        \label{fig:jaccardFire_assimilation}
    \end{subfigure}
    \begin{subfigure}[t]{3.2in}
        \includegraphics[width=3.2in]{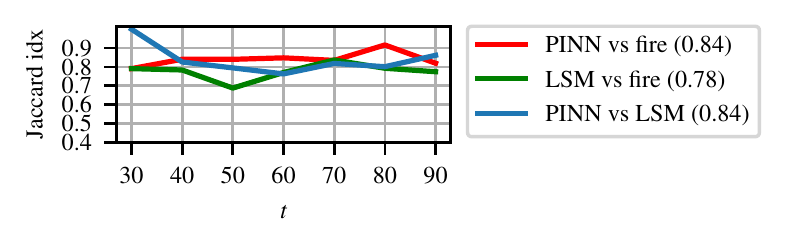}
        \caption{Fire E06: PINN-a versus the level-set method.}
        \label{fig:jaccardFire_assimilation_E06_Braidwood}
    \end{subfigure}
    \begin{subfigure}[t]{3.2in}
        \includegraphics[width=3.2in]{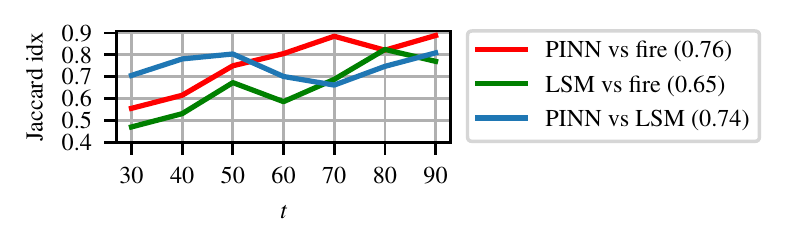}
        \caption{Fire S03: B-PINN-a versus the level-set method.}
        \label{fig:jaccardFire_assimilation_bpinn}
    \end{subfigure}
    \begin{subfigure}[t]{3.2in}
        \includegraphics[width=3.2in]{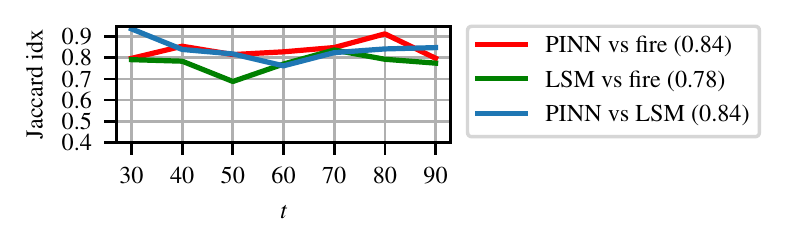}
        \caption{Fire E06: B-PINN-a versus the level-set method.}
        \label{fig:jaccardFire_assimilation_bpinn_E06_Braidwood}
    \end{subfigure}
    \caption{Jaccard index results for fire S03 (left-hand column) and fire E06 (right-hand column). LSM denotes the level-set method. Mean values across time are provided in brackets in the legend.}
    \label{fig:jaccardFire}
\end{figure}

The Jaccard index for the PINN-f and level-set method are plotted in Figure \ref{fig:jaccardFire_noAssimilation}. The PINN-f and level-set method predictions are compared with each other, as well as with ground-truth fire data. The results indicate that both the PINN-f and level-set methods are initially not accurate, but improve over time. The initial lack of accuracy is due to an over-estimation of the initial fire front ellipse angle. The Jaccard index between the PINN-f and level-set method remains above $70\%$, indicating that both methods produce similar prediction of the fire-front. However, the level-set method generally has a slightly higher Jaccard indices than the PINN-f, suggesting that it performs marginally better.

%

%

\begin{figure}[!t]
    \centering
    \begin{subfigure}[t]{3.2in}
        \includegraphics[width=3.2in]{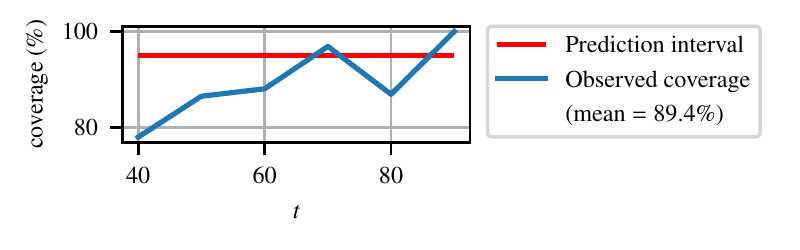}
        \caption{Fire S03: uncertainty coverage.}
        \label{fig:coverage_S03_Braidwood}
    \end{subfigure}
    \begin{subfigure}[t]{3.2in}
        \includegraphics[width=3.2in]{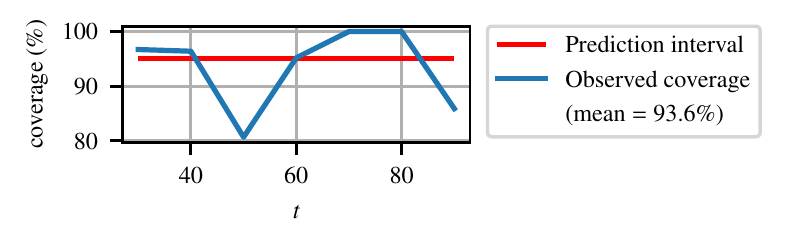}
        \caption{Fire E06: uncertainty coverage.}
        \label{fig:coverage_E06_Braidwood}
    \end{subfigure}
    \caption{Uncertainty coverage of the B-PINN-a model for fires S03 (left figure) and E06 (right figure). The mean values indicated in the legends are computed over time.}
\end{figure}

Plots of the fire-front predictions for the PINN-a and the level-set method are shown in Figure \ref{fig:fireResults_assimilation}. The PINN-a results follow actual fire more closely due to the assimilation of the fire-front isochrone data into the model. It initially fits to the elliptical cone of $\mathcal{D}_i$ and transitions closer to the fire-front isochrones over time. The PINN does not over-fit to these isochrones as it is regularised by the limited size of the neural network and by early stopping. 

The Jaccard index results for the PINN-a and level-set method are plotted in Figure \ref{fig:jaccardFire_assimilation}. The Jaccard indices have increased from 0.65 for the level-set method to 0.74 for the PINN-a indicating that the PINN-a provides a better simulation of the fire. Furthermore, the average value of the Jaccard indices between the PINN-a and the level-set method over time is $82\%$, indicating that the PINN-a still maintains the physics defined by the level-set equation.

The B-PINN-a and the level-set method's fire-front predictions are presented in Figure \ref{fig:fireResults_assimilation_bpinn}. To quantify uncertainty, a set of 100 MC samples of B-PINN-a parameters are drawn from the posterior distribution. These are used to produce a set of 100 MC simulations of fire-front predictions, each corresponding to one of the 100 MC parameters. The 95\% posterior predictive interval is computed from these 100 MC predictions and is plotted by the grey region in Figure \ref{fig:fireResults_assimilation_bpinn} and the mean of the MC simulations is illustrated by the red isochrone in the plot. Note that the MC simulations demonstrate variation in both the size and the shape of the isochrones. This is indicated by the variation in the shape of the mean and 95\% posterior predictive interval plots. Compared to the PINN-f, the data assimilation in B-PINN-a allows it to produce results that are closer to the data.

The Jaccard index for the B-PINN-a results are shown in Figure \ref{fig:jaccardFire_assimilation_bpinn}. We find that, as for the PINN-a, the data assimilation draws the predictions closer to the data to produce a more accurate representation of the fire compared to the level-set method. The average Jaccard index for the B-PINN-a is higher than the PINN-a, however the increase is marginal. However, if the PINN-a is not regularised in any way, the B-PINN-a has the potential to provide more accurate predictions due to the natural regularisation introduced by the prior.

The coverage results for the B-PINN-a are illustrated in Figure \ref{fig:coverage_S03_Braidwood}. Owing to the over estimation of the fire-front at $t=30$, the coverage for this first result is excluded. The average coverage over time is 89.4\%, which is close to the 95\% target. This indicates a slight underestimation of the the uncertainty, which as described in section \ref{sec:bpinn}, could be due to the mean-field assumption. This result could be improved by reducing the mean-field independence assumptions or by using a Markov Chain Monte Carlo inference approach such as Hamilton Monte Carlo \cite{neal2011mcmc,yang2021bpinns}.

\subsubsection{Fire E06: Dominant Wind Fire Dynamics}

\begin{figure}[!p]
    \centering
    \begin{subfigure}[t]{5.39in}
        \includegraphics[width=5.39in]{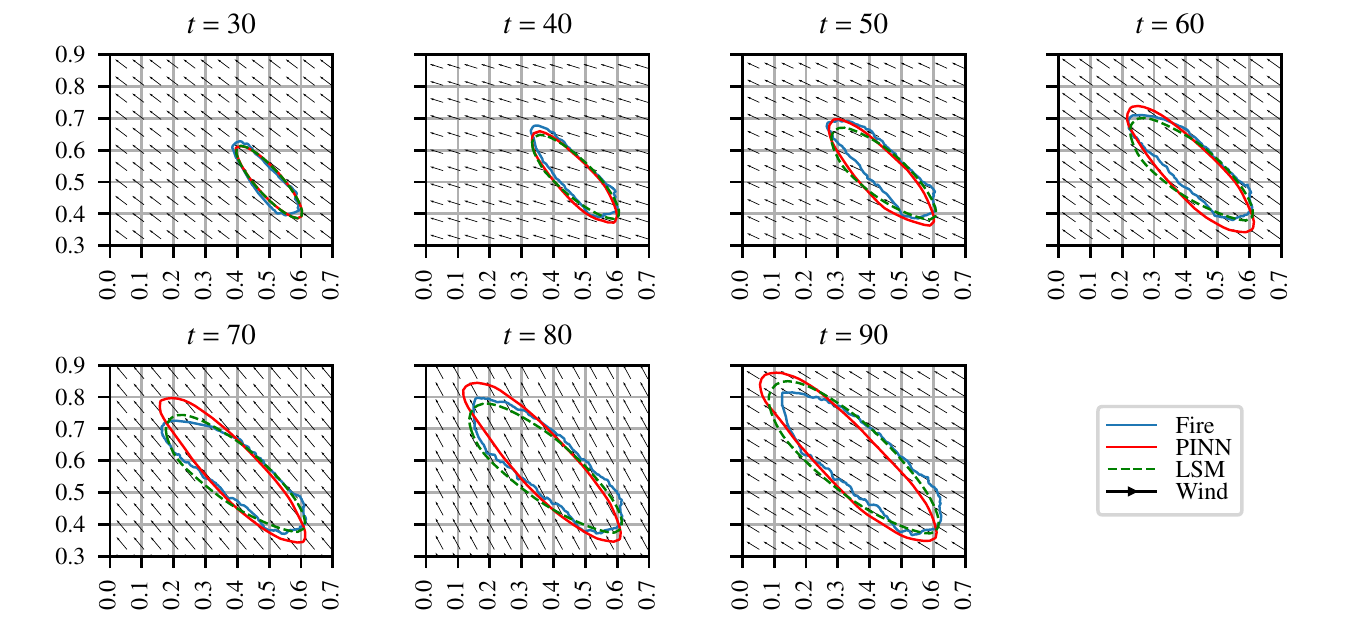}
        \caption{PINN-f versus the level-set method.}
        \label{fig:fireResults_noAssimilation_E06_Braidwood}
    \end{subfigure}
    \begin{subfigure}[t]{5.39in}
        \includegraphics[width=5.39in]{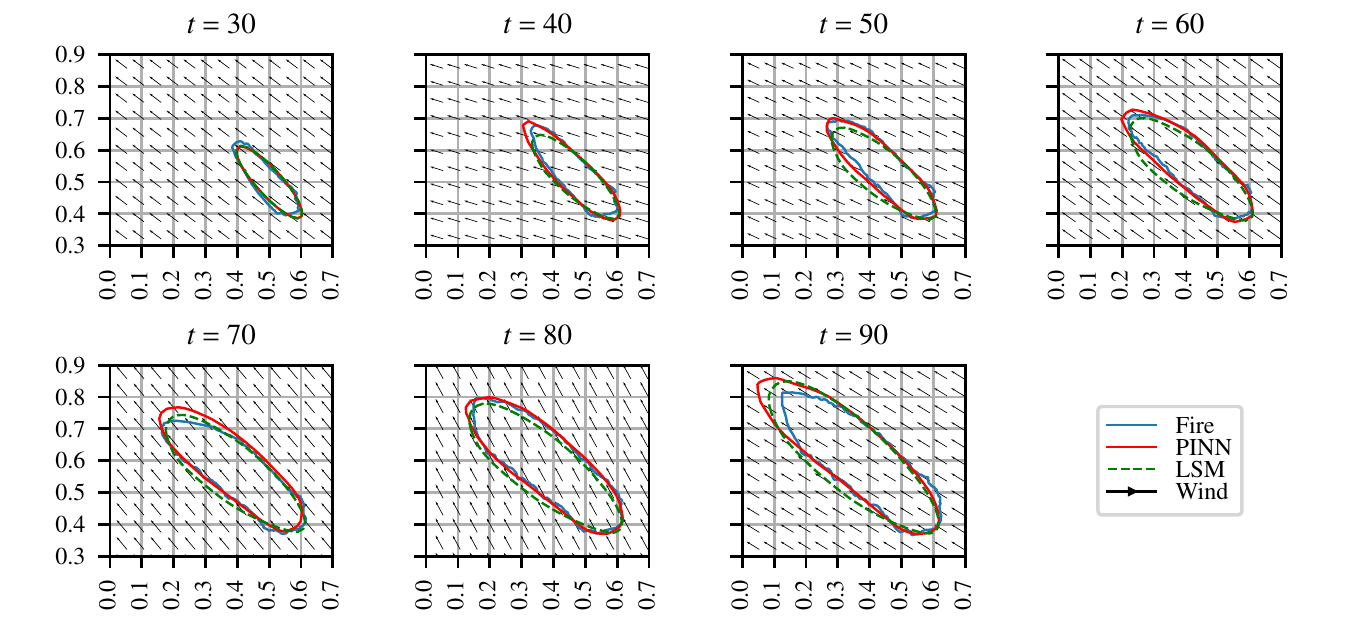}
        \caption{PINN-a versus the level-set method.}
        \label{fig:fireResults_assimilation_E06_Braidwood}
    \end{subfigure}
    \begin{subfigure}[t]{5.39in}
        \includegraphics[width=5.39in]{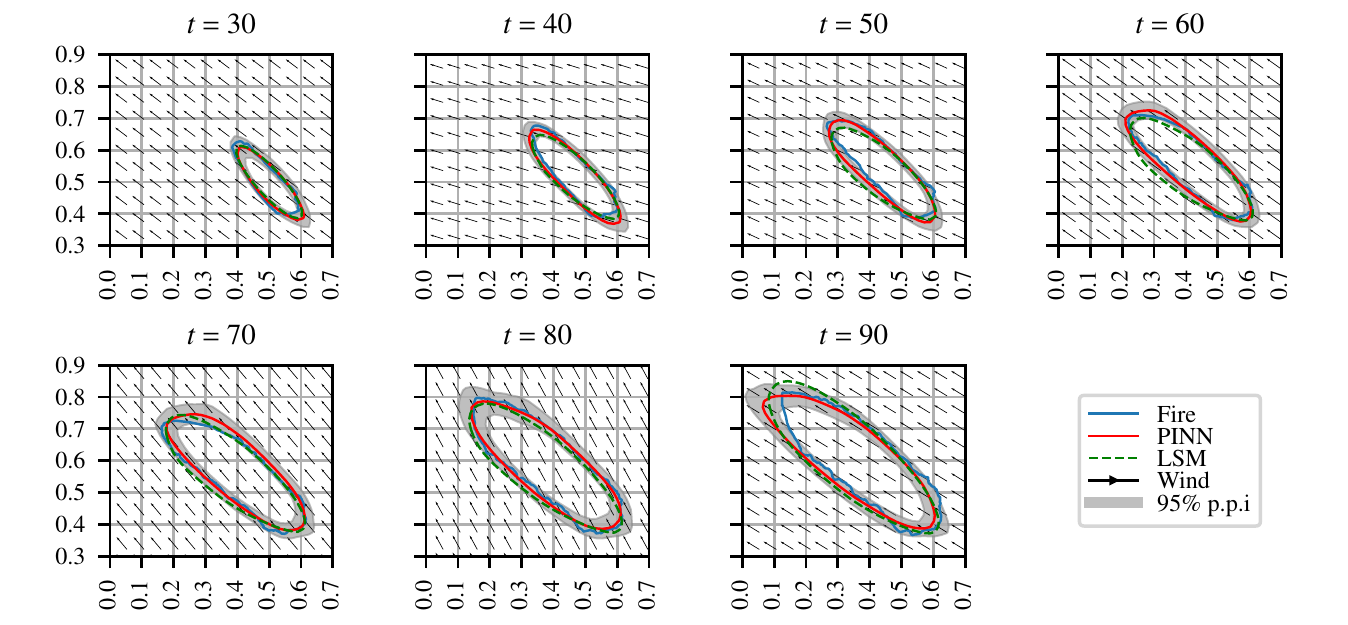}
        \caption{B-PINN-a versus the level-set method.}
        \label{fig:fireResults_assimilation_bpinn_E06_Braidwood}
    \end{subfigure}
    \caption{PINN and Level-Set Method results on the E06 fire dataset. LSM denotes the level-set method and p.p.i. denotes posterior predictive interval.}
    \label{fig:fireResults_E06_Braidwood}
\end{figure}

The E06 fire simulations for the PINN-f, PINN-a, B-PINN-a, and level-set method are provided in Figure \ref{fig:fireResults_E06_Braidwood} and the Jaccard index plots are provides in the right-hand column of Figure \ref{fig:jaccardFire}. Compared to fire S03, fire E06 does not accelerate at the beginning stages of its burning, and the predictions for this fire are thus more accurate. The PINN-f propagates the fire-front slightly faster than the level-set method. Data assimilation in the PINN-a constrains the fire closer to the data to increase the average Jaccard index from 0.72 to 0.84. The B-PINN-a provides a distribution of MC simulations of the fire providing uncertainty quantification. The variance of the distribution tends to be higher in the directions where the fire moves at higher speeds, which is where the spatial gradient of the zero-level-set is high and is in the direction that the wind is blowing. According to the Jaccard indices, the B-PINN-a and the PINN-a provide similar accuracy. The B-PINN-a however additionally offers uncertainty quantification. The coverage results for the B-PINN-a are illustrated in Figure \ref{fig:coverage_E06_Braidwood}. The average coverage over time is 93.6\%, which is very close to the 95\% target. 

\section{Discussion}

Compared to traditional linear solvers (such as the level-set method), the PINN offers several appealing properties. These include a functional-form solution over time and space, a non-linear representation of the solution, the ability to perform data assimilation, and the ability to provide uncertainty quantification. The level-set method uses linear approximations through a discretisation of space and time, which is often required to be at a high resolution. Furthermore, as the level-set function evolves through time, large variations in the gradients may develop, requiring a reinitialisation scheme to constrain the level-set function. The PINN does not require discretisation and naturally handles non-linearity.

Data assimilation is a key benefit of our approach as it allows for the correction of errors in the propagation of the level-set function over time. Furthermore, it provides a means to perform post-fire reconstruction where the PINN can be used to interpolate between two or more observed fire isochrones to understand how a fire may have evolved. Such observed fire isochrones may be obtained with existing automated methods that process remote sensed data (e.g., \cite{Schroeder2014New}). Furthermore, the data assimilated into the model may originate from multiple sources such as on-the-ground observations of the fire-front (e.g., by fire fighters), data captured by drones, or remote sensed data such as satellites.

Finally, the B-PINN provides a means for uncertainty quantification, which is not naturally available in the level-set method. Though uncertainty quantification is the main benefit of the B-PINN, the B-PINN also offers a means to produce various plausible fire-front simulations via MC samples. This is useful for decision makers to understand the fire dynamics and is also useful for post-fire reconstruction.

When performing Bayesian inference in neural networks three challenges typically arise: (i) due to the ``black box'' nature of these models, specifying a meaningful prior is difficult; (ii) the model parameters (e.g., weights and biases) are usually non-identifiable (e.g., see \cite{mcdermott2019bayesian}); and (iii) owing to the over-parametrisation of the model, Bayesian inference can be computationally expensive. MCMC methods (such as HMC) would provide an accurate uncertainty quantification, however these methods become prohibitively slow and expensive when the models are large. Variational inference offers a fast alternative with its gradient descent based optimisation approach. Furthermore, though this optimisation approach and the mean-field assumptions can be viewed as limiting the functional form of the posterior distribution, this limitation can also be viewed as a form of regularisation, forcing the model to express the posterior simply and may help to alleviate the identifiability of neural network parameters from data.

Relating to PINN challenges, we encountered problems associated with spectral bias and temporal continuity. Regarding spectral bias, the PINN was slow to learn the high-frequency features associated with the discontinuity of the signed distance function. To address this, we over-sampled the region within the zero-level-set and we trained the model for a large number of epochs. Owing to its piece-wise nature, the Rectified Linear Unit (ReLU) activation function provides better approximations to the signed-distance function. This however comes at the cost of introducing discontinuities into the gradients. For the synthetic dataset, the tanh activation function was used, and for the fire dataset, the ReLU activation function was used.

The temporal continuity problem in the PINN was a significant challenge encountered in this study. Strategies such as sequence-to-sequence learning \cite{krishnapriyan2021characterizing} and weighted residual cost \cite{wang2022respecting} were not effective in our problem and led to the development of the proposed forecast likelihood. Introducing this likelihood provided significant improvement and allowed the PINN to model changes in the external wind vector field. We believe that this approach provides a general solution that will be effective in other applications too.

Future work may include investigating the proposed approach on datasets comprising ``mega-fires''. Mega-fires are typically characterised as fires that have an unusually large size for a given ecosystem \cite{pausas2021wildfires}. With wide coverage, these fires often evolve over non-homogenous environments. Furthermore, large fires can generate their own wind as the plume draws air in. This can significantly affect the wind conditions and spot fires developing downwind of the plume can move backwards against the expected wind direction. Such effects are not taken into account in simple models such as the Rothermel model, but they can be incorporated with various parametrisations of the surface pressure effects \cite{hilton2018incorporating}.

\section{Conclusion}

We contrast and compare the PINN, the B-PINN, and the level-set method in solving the level-set equation in the context of wildfire fire-front modelling. We propose a novel approach to address the temporal continuity problem in PINNs with a forecast likelihood and demonstrate its importance and effectiveness through a synthetic dataset. Furthermore, we develop a novel approach to perform data assimilation in the PINN: a capability that the level-set method does not naturally offer. With the proposed approaches, the PINN and B-PINN are successfully applied to modelling two recorded grassland fires where we demonstrate data assimilation and uncertainty quantification. These results are especially significant given the challenges associated with optimising PINNs in real-world applications and the fact that the level-set method does not naturally offer the capabilities demonstrated. We thus argue that the PINN provides a practically useful solver for the level-set function and could be integrated into existing wildfire modelling platforms to facilitate data assimilation and uncertainty quantification.

\section*{Acknowledgements}
This work was supported by the CSIRO MLAI Future Science Platform.



  \bibliographystyle{elsarticle-num-names} 
  \bibliography{Bibliography}

\begin{thebibliography}{70}
\expandafter\ifx\csname natexlab\endcsname\relax\def\natexlab#1{#1}\fi
\providecommand{\url}[1]{\texttt{#1}}
\providecommand{\href}[2]{#2}
\providecommand{\path}[1]{#1}
\providecommand{\DOIprefix}{doi:}
\providecommand{\ArXivprefix}{arXiv:}
\providecommand{\URLprefix}{URL: }
\providecommand{\Pubmedprefix}{pmid:}
\providecommand{\doi}[1]{\href{http://dx.doi.org/#1}{\path{#1}}}
\providecommand{\Pubmed}[1]{\href{pmid:#1}{\path{#1}}}
\providecommand{\bibinfo}[2]{#2}
\ifx\xfnm\relax \def\xfnm[#1]{\unskip,\space#1}\fi
\bibitem[{Australia(2022)}]{wwf2022australia}
\bibinfo{author}{W.~Australia}, \bibinfo{title}{Australia bushfires: Recovering
  and restoring australia after the devastating 2019-20 bushfires.},
  \bibinfo{howpublished}{"online:
  \url{https://www.wwf.org.au/what-we-do/bushfires\#gs.i2xrqm}"},
  \bibinfo{year}{2022}. \bibinfo{note}{Last accessed Nov 2022.}
\bibitem[{Bowman et~al.(2020)Bowman, Kolden, Abatzoglou, Johnston, van~der
  Werf, and Flannigan}]{Bowman_2020}
\bibinfo{author}{D.~M. J.~S. Bowman}, \bibinfo{author}{C.~A. Kolden},
  \bibinfo{author}{J.~T. Abatzoglou}, \bibinfo{author}{F.~H. Johnston},
  \bibinfo{author}{G.~R. van~der Werf}, \bibinfo{author}{M.~Flannigan},
\newblock \bibinfo{title}{Vegetation fires in the anthropocene},
\newblock \bibinfo{journal}{Nature Reviews Earth \& Environment}
  \bibinfo{volume}{1} (\bibinfo{year}{2020}) \bibinfo{pages}{500--515}.
\bibitem[{Huston et~al.(2022)Huston, Davis, Kuhnert, and
  Bolt}]{huston2022creating}
\bibinfo{author}{C.~Huston}, \bibinfo{author}{J.~Davis},
  \bibinfo{author}{P.~Kuhnert}, \bibinfo{author}{A.~Bolt},
\newblock \bibinfo{title}{Creating trusted extensions to existing software
  tools in bushfire consequence estimation},
\newblock in: \bibinfo{booktitle}{ISCRAM Asia Pacific 2022, Dealing with the
  Unexpected}, \bibinfo{address}{Melbourne Australia}, \bibinfo{year}{2022}.
\bibitem[{Miller et~al.(2015)Miller, Hilton, Sullivan, and
  Prakash}]{miller2015spark}
\bibinfo{author}{C.~Miller}, \bibinfo{author}{J.~Hilton},
  \bibinfo{author}{A.~Sullivan}, \bibinfo{author}{M.~Prakash},
\newblock \bibinfo{title}{Spark -- a bushfire spread prediction tool},
\newblock in: \bibinfo{editor}{R.~Denzer}, \bibinfo{editor}{R.~M. Argent},
  \bibinfo{editor}{G.~Schimak}, \bibinfo{editor}{J.~Hreb{\'i}cek} (Eds.),
  \bibinfo{booktitle}{Environmental Software Systems. Infrastructures, Services
  and Applications}, \bibinfo{publisher}{Springer International Publishing},
  \bibinfo{address}{Cham}, \bibinfo{year}{2015}, pp. \bibinfo{pages}{262--271}.
\bibitem[{Mandel et~al.(2011)Mandel, Beezley, and
  Kochanski}]{mandel2011coupled}
\bibinfo{author}{J.~Mandel}, \bibinfo{author}{J.~D. Beezley},
  \bibinfo{author}{A.~K. Kochanski},
\newblock \bibinfo{title}{Coupled atmosphere-wildland fire modeling with wrf
  3.3 and sfire 2011},
\newblock \bibinfo{journal}{Geoscientific Model Development}
  \bibinfo{volume}{4} (\bibinfo{year}{2011}) \bibinfo{pages}{591--610}.
\bibitem[{Osher and Fedkiw(2001)}]{Osher2001Level}
\bibinfo{author}{S.~Osher}, \bibinfo{author}{R.~P. Fedkiw},
\newblock \bibinfo{title}{Level set methods: An overview and some recent
  results},
\newblock \bibinfo{journal}{Journal of Computational Physics}
  \bibinfo{volume}{169} (\bibinfo{year}{2001}) \bibinfo{pages}{463--502}.
  \URLprefix
  \url{https://www.sciencedirect.com/science/article/pii/S0021999100966361}.
  \DOIprefix\doi{https://doi.org/10.1006/jcph.2000.6636}.
\bibitem[{Sethian(1999)}]{sethian1999level}
\bibinfo{author}{J.~Sethian}, \bibinfo{title}{Level Set Methods and Fast
  Marching Methods: Evolving Interfaces in Computational Geometry, Fluid
  Mechanics, Computer Vision, and Materials Science}, Cambridge Monographs on
  Applied and Computational Mathematics, \bibinfo{publisher}{Cambridge
  University Press}, \bibinfo{year}{1999}.
\bibitem[{Yoo and Wikle(2022)}]{yoo2022bayesian}
\bibinfo{author}{M.~Yoo}, \bibinfo{author}{C.~K. Wikle},
\newblock \bibinfo{title}{A bayesian spatio-temporal level set dynamic model
  and application to fire front propagation},
\newblock \bibinfo{journal}{arXiv preprint arXiv:2210.14978}
  (\bibinfo{year}{2022}).
\bibitem[{Dabrowski et~al.(2022)Dabrowski, Huston, Hilton, Mangeon, and
  Kuhnert}]{dabrowski2022towards}
\bibinfo{author}{J.~J. Dabrowski}, \bibinfo{author}{C.~Huston},
  \bibinfo{author}{J.~Hilton}, \bibinfo{author}{S.~Mangeon},
  \bibinfo{author}{P.~Kuhnert},
\newblock \bibinfo{title}{Towards data assimilation in level-set wildfire
  models using bayesian filtering},
\newblock \bibinfo{journal}{arXiv preprint arXiv:2206.08501}
  (\bibinfo{year}{2022}).
\bibitem[{Kuhnert et~al.(2018)Kuhnert, Pagendam, Bartley, Gladish, Lewis, and
  Bainbridge}]{kuhnert2018making}
\bibinfo{author}{P.~Kuhnert}, \bibinfo{author}{D.~Pagendam},
  \bibinfo{author}{R.~Bartley}, \bibinfo{author}{D.~Gladish},
  \bibinfo{author}{S.~Lewis}, \bibinfo{author}{Z.~Bainbridge},
\newblock \bibinfo{title}{Making management decisions in the face of
  uncertainty: a case study using the burdekin catchment in the great barrier
  reef},
\newblock \bibinfo{journal}{Marine and Freshwater Research}
  \bibinfo{volume}{69} (\bibinfo{year}{2018}) \bibinfo{pages}{1187--1200}.
\bibitem[{Rochoux et~al.(2013)Rochoux, Delmotte, Cuenot, Ricci, and
  Trouvé}]{Rochoux2013Regional}
\bibinfo{author}{M.~Rochoux}, \bibinfo{author}{B.~Delmotte},
  \bibinfo{author}{B.~Cuenot}, \bibinfo{author}{S.~Ricci},
  \bibinfo{author}{A.~Trouvé},
\newblock \bibinfo{title}{Regional-scale simulations of wildland fire spread
  informed by real-time flame front observations},
\newblock \bibinfo{journal}{Proceedings of the Combustion Institute}
  \bibinfo{volume}{34} (\bibinfo{year}{2013}) \bibinfo{pages}{2641--2647}.
  \URLprefix
  \url{https://www.sciencedirect.com/science/article/pii/S1540748912001988}.
  \DOIprefix\doi{https://doi.org/10.1016/j.proci.2012.06.090}.
\bibitem[{Lagaris et~al.(1998)Lagaris, Likas, and
  Fotiadis}]{Lagaris1998Artificial}
\bibinfo{author}{I.~Lagaris}, \bibinfo{author}{A.~Likas},
  \bibinfo{author}{D.~Fotiadis},
\newblock \bibinfo{title}{Artificial neural networks for solving ordinary and
  partial differential equations},
\newblock \bibinfo{journal}{IEEE Transactions on Neural Networks}
  \bibinfo{volume}{9} (\bibinfo{year}{1998}) \bibinfo{pages}{987--1000}.
  \DOIprefix\doi{10.1109/72.712178}.
\bibitem[{Raissi et~al.(2019)Raissi, Perdikaris, and
  Karniadakis}]{raissi2019physics}
\bibinfo{author}{M.~Raissi}, \bibinfo{author}{P.~Perdikaris},
  \bibinfo{author}{G.~Karniadakis},
\newblock \bibinfo{title}{Physics-informed neural networks: A deep learning
  framework for solving forward and inverse problems involving nonlinear
  partial differential equations},
\newblock \bibinfo{journal}{Journal of Computational Physics}
  \bibinfo{volume}{378} (\bibinfo{year}{2019}) \bibinfo{pages}{686--707}.
  \URLprefix
  \url{https://www.sciencedirect.com/science/article/pii/S0021999118307125}.
  \DOIprefix\doi{https://doi.org/10.1016/j.jcp.2018.10.045}.
\bibitem[{Cuomo et~al.(2022)Cuomo, Di~Cola, Giampaolo, Rozza, Raissi, and
  Piccialli}]{Cuomo2022Scientific}
\bibinfo{author}{S.~Cuomo}, \bibinfo{author}{V.~S. Di~Cola},
  \bibinfo{author}{F.~Giampaolo}, \bibinfo{author}{G.~Rozza},
  \bibinfo{author}{M.~Raissi}, \bibinfo{author}{F.~Piccialli},
\newblock \bibinfo{title}{Scientific machine learning through physics--informed
  neural networks: Where we are and what's next},
\newblock \bibinfo{journal}{Journal of Scientific Computing}
  \bibinfo{volume}{92} (\bibinfo{year}{2022}) \bibinfo{pages}{88}. \URLprefix
  \url{https://doi.org/10.1007/s10915-022-01939-z}.
  \DOIprefix\doi{10.1007/s10915-022-01939-z}.
\bibitem[{Markidis(2021)}]{markidis2021old}
\bibinfo{author}{S.~Markidis},
\newblock \bibinfo{title}{The old and the new: Can {Physics-Informed}
  {Deep-Learning} replace traditional linear solvers?},
\newblock \bibinfo{journal}{Front Big Data} \bibinfo{volume}{4}
  (\bibinfo{year}{2021}) \bibinfo{pages}{669097}.
\bibitem[{He et~al.(2020)He, Barajas-Solano, Tartakovsky, and
  Tartakovsky}]{he2020physics}
\bibinfo{author}{Q.~He}, \bibinfo{author}{D.~Barajas-Solano},
  \bibinfo{author}{G.~Tartakovsky}, \bibinfo{author}{A.~M. Tartakovsky},
\newblock \bibinfo{title}{Physics-informed neural networks for multiphysics
  data assimilation with application to subsurface transport},
\newblock \bibinfo{journal}{Advances in Water Resources} \bibinfo{volume}{141}
  (\bibinfo{year}{2020}) \bibinfo{pages}{103610}. \URLprefix
  \url{https://www.sciencedirect.com/science/article/pii/S0309170819311649}.
  \DOIprefix\doi{https://doi.org/10.1016/j.advwatres.2020.103610}.
\bibitem[{Yang et~al.(2021)Yang, Meng, and Karniadakis}]{yang2021bpinns}
\bibinfo{author}{L.~Yang}, \bibinfo{author}{X.~Meng}, \bibinfo{author}{G.~E.
  Karniadakis},
\newblock \bibinfo{title}{B-pinns: Bayesian physics-informed neural networks
  for forward and inverse pde problems with noisy data},
\newblock \bibinfo{journal}{Journal of Computational Physics}
  \bibinfo{volume}{425} (\bibinfo{year}{2021}) \bibinfo{pages}{109913}.
  \URLprefix
  \url{https://www.sciencedirect.com/science/article/pii/S0021999120306872}.
  \DOIprefix\doi{https://doi.org/10.1016/j.jcp.2020.109913}.
\bibitem[{Wang et~al.(2022)Wang, Sankaran, and Perdikaris}]{wang2022respecting}
\bibinfo{author}{S.~Wang}, \bibinfo{author}{S.~Sankaran},
  \bibinfo{author}{P.~Perdikaris},
\newblock \bibinfo{title}{Respecting causality is all you need for training
  physics-informed neural networks},
\newblock \bibinfo{journal}{arXiv preprint arXiv:2203.07404}
  (\bibinfo{year}{2022}).
\bibitem[{Mojgani et~al.(2022)Mojgani, Balajewicz, and
  Hassanzadeh}]{mojgani2022lagrangian}
\bibinfo{author}{R.~Mojgani}, \bibinfo{author}{M.~Balajewicz},
  \bibinfo{author}{P.~Hassanzadeh},
\newblock \bibinfo{title}{Lagrangian pinns: A causality-conforming solution to
  failure modes of physics-informed neural networks},
\newblock \bibinfo{journal}{arXiv preprint arXiv:2205.02902}
  (\bibinfo{year}{2022}).
\bibitem[{Kuhnert et~al.(2010)Kuhnert, Henderson, Bartley, and
  Herr}]{kuhnert2010incorporating}
\bibinfo{author}{P.~M. Kuhnert}, \bibinfo{author}{A.-K. Henderson},
  \bibinfo{author}{R.~Bartley}, \bibinfo{author}{A.~Herr},
\newblock \bibinfo{title}{Incorporating uncertainty in gully erosion
  calculations using the random forests modelling approach},
\newblock \bibinfo{journal}{Environmetrics} \bibinfo{volume}{21}
  (\bibinfo{year}{2010}) \bibinfo{pages}{493--509}.
\bibitem[{Zammit-Mangion and Cressie(2021)}]{ZammitMangion2021frk}
\bibinfo{author}{A.~Zammit-Mangion}, \bibinfo{author}{N.~Cressie},
\newblock \bibinfo{title}{Frk: An r package for spatial and spatio-temporal
  prediction with large datasets},
\newblock \bibinfo{journal}{Journal of Statistical Software}
  \bibinfo{volume}{98} (\bibinfo{year}{2021}) \bibinfo{pages}{1–48}.
  \URLprefix
  \url{https://www.jstatsoft.org/index.php/jss/article/view/v098i04}.
  \DOIprefix\doi{10.18637/jss.v098.i04}.
\bibitem[{Kuhnert(2014)}]{kuhnert2014physical}
\bibinfo{author}{P.~M. Kuhnert},
\newblock \bibinfo{title}{Physical-statistical modeling},
\newblock \bibinfo{journal}{Wiley StatsRef: Statistics Reference Online}
  (\bibinfo{year}{2014}) \bibinfo{pages}{1--5}.
\bibitem[{Wikle et~al.(1998)Wikle, Berliner, and
  Cressie}]{wikle1998hierarchical}
\bibinfo{author}{C.~K. Wikle}, \bibinfo{author}{L.~M. Berliner},
  \bibinfo{author}{N.~Cressie},
\newblock \bibinfo{title}{Hierarchical bayesian space-time models},
\newblock \bibinfo{journal}{Environmental and ecological statistics}
  \bibinfo{volume}{5} (\bibinfo{year}{1998}) \bibinfo{pages}{117--154}.
\bibitem[{Wikle and Hooten(2010)}]{wikle2010general}
\bibinfo{author}{C.~K. Wikle}, \bibinfo{author}{M.~B. Hooten},
\newblock \bibinfo{title}{A general science-based framework for dynamical
  spatio-temporal models},
\newblock \bibinfo{journal}{Test} \bibinfo{volume}{19} (\bibinfo{year}{2010})
  \bibinfo{pages}{417--451}.
\bibitem[{Gladish et~al.(2016)Gladish, Lewis, Bainbridge, Brodie, Kuhnert,
  Pagendam, Wikle, Bartley, Searle, Ellis, Dougall, and
  Turner}]{gladish2016spatiotemporal}
\bibinfo{author}{D.~W. Gladish}, \bibinfo{author}{S.~E. Lewis},
  \bibinfo{author}{Z.~T. Bainbridge}, \bibinfo{author}{J.~E. Brodie},
  \bibinfo{author}{P.~M. Kuhnert}, \bibinfo{author}{D.~E. Pagendam},
  \bibinfo{author}{C.~K. Wikle}, \bibinfo{author}{R.~Bartley},
  \bibinfo{author}{R.~D. Searle}, \bibinfo{author}{R.~J. Ellis},
  \bibinfo{author}{C.~Dougall}, \bibinfo{author}{R.~D.~R. Turner},
\newblock \bibinfo{title}{{Spatio-temporal assimilation of modelled catchment
  loads with monitoring data in the Great Barrier Reef}},
\newblock \bibinfo{journal}{The Annals of Applied Statistics}
  \bibinfo{volume}{10} (\bibinfo{year}{2016}) \bibinfo{pages}{1590 -- 1618}.
  \URLprefix \url{https://doi.org/10.1214/16-AOAS950}.
  \DOIprefix\doi{10.1214/16-AOAS950}.
\bibitem[{Wikle et~al.(2019)Wikle, Zammit-Mangion, and
  Cressie}]{wikle2019spatio}
\bibinfo{author}{C.~K. Wikle}, \bibinfo{author}{A.~Zammit-Mangion},
  \bibinfo{author}{N.~Cressie}, \bibinfo{title}{Spatio-temporal Statistics with
  R}, \bibinfo{publisher}{Chapman and Hall/CRC}, \bibinfo{year}{2019}.
\bibitem[{Bolt et~al.(2022)Bolt, Huston, Kuhnert, Dabrowski, Hilton, and
  Sanderson}]{bolt2022spatiotemporal}
\bibinfo{author}{A.~Bolt}, \bibinfo{author}{C.~Huston},
  \bibinfo{author}{P.~Kuhnert}, \bibinfo{author}{J.~J. Dabrowski},
  \bibinfo{author}{J.~Hilton}, \bibinfo{author}{C.~Sanderson},
\newblock \bibinfo{title}{A spatio-temporal neural network forecasting approach
  for emulation of firefront models},
\newblock in: \bibinfo{booktitle}{2022 Signal Processing: Algorithms,
  Architectures, Arrangements, and Applications (SPA)}, \bibinfo{year}{2022},
  pp. \bibinfo{pages}{110--115}.
  \DOIprefix\doi{10.23919/SPA53010.2022.9927888}.
\bibitem[{Mallet et~al.(2009)Mallet, Keyes, and Fendell}]{mallet2009modeling}
\bibinfo{author}{V.~Mallet}, \bibinfo{author}{D.~Keyes},
  \bibinfo{author}{F.~Fendell},
\newblock \bibinfo{title}{Modeling wildland fire propagation with level set
  methods},
\newblock \bibinfo{journal}{Computers \& Mathematics with Applications}
  \bibinfo{volume}{57} (\bibinfo{year}{2009}) \bibinfo{pages}{1089--1101}.
  \URLprefix
  \url{https://www.sciencedirect.com/science/article/pii/S0898122108006329}.
  \DOIprefix\doi{https://doi.org/10.1016/j.camwa.2008.10.089}.
\bibitem[{Hilton et~al.(2015)Hilton, Miller, Sullivan, and
  Rucinski}]{hilton2015effects}
\bibinfo{author}{J.~Hilton}, \bibinfo{author}{C.~Miller},
  \bibinfo{author}{A.~Sullivan}, \bibinfo{author}{C.~Rucinski},
\newblock \bibinfo{title}{Effects of spatial and temporal variation in
  environmental conditions on simulation of wildfire spread},
\newblock \bibinfo{journal}{Environmental Modelling \& Software}
  \bibinfo{volume}{67} (\bibinfo{year}{2015}) \bibinfo{pages}{118--127}.
  \URLprefix
  \url{https://www.sciencedirect.com/science/article/pii/S1364815215000468}.
  \DOIprefix\doi{https://doi.org/10.1016/j.envsoft.2015.01.015}.
\bibitem[{Hilton et~al.(2016)Hilton, Miller, Sharples, and
  Sullivan}]{hilton2016curvature}
\bibinfo{author}{J.~Hilton}, \bibinfo{author}{C.~Miller},
  \bibinfo{author}{J.~Sharples}, \bibinfo{author}{A.~Sullivan},
\newblock \bibinfo{title}{Curvature effects in the dynamic propagation of
  wildfires},
\newblock \bibinfo{journal}{International Journal of Wildland Fire}
  \bibinfo{volume}{25} (\bibinfo{year}{2016}) \bibinfo{pages}{1238--1251}.
\bibitem[{Osher and Fedkiw(2003)}]{osher2003level}
\bibinfo{author}{S.~Osher}, \bibinfo{author}{R.~P. Fedkiw},
  \bibinfo{title}{Level set methods and dynamic implicit surfaces}, volume
  \bibinfo{volume}{153}, \bibinfo{publisher}{Springer}, \bibinfo{year}{2003}.
\bibitem[{Srivas et~al.(2016)Srivas, Artés, {de Callafon}, and
  Altintas}]{Srivas2016Wildfire}
\bibinfo{author}{T.~Srivas}, \bibinfo{author}{T.~Artés},
  \bibinfo{author}{R.~A. {de Callafon}}, \bibinfo{author}{I.~Altintas},
\newblock \bibinfo{title}{Wildfire spread prediction and assimilation for
  farsite using ensemble kalman filtering},
\newblock \bibinfo{journal}{Procedia Computer Science} \bibinfo{volume}{80}
  (\bibinfo{year}{2016}) \bibinfo{pages}{897--908}. \URLprefix
  \url{https://www.sciencedirect.com/science/article/pii/S187705091630727X}.
  \DOIprefix\doi{https://doi.org/10.1016/j.procs.2016.05.328},
  \bibinfo{note}{international Conference on Computational Science 2016, ICCS
  2016, 6-8 June 2016, San Diego, California, USA}.
\bibitem[{Xue et~al.(2012)Xue, Gu, and Hu}]{Xue2012Data}
\bibinfo{author}{H.~Xue}, \bibinfo{author}{F.~Gu}, \bibinfo{author}{X.~Hu},
\newblock \bibinfo{title}{Data assimilation using sequential monte carlo
  methods in wildfire spread simulation},
\newblock \bibinfo{journal}{ACM Trans. Model. Comput. Simul.}
  \bibinfo{volume}{22} (\bibinfo{year}{2012}). \URLprefix
  \url{https://doi.org/10.1145/2379810.2379816}.
  \DOIprefix\doi{10.1145/2379810.2379816}.
\bibitem[{Silva et~al.(2014)Silva, Rochoux, Orlande, Colaco, Fudym, El-Hafi,
  Cuenot, and Ricci}]{Silva2014Application}
\bibinfo{author}{W.~B.~d. Silva}, \bibinfo{author}{M.~C. Rochoux},
  \bibinfo{author}{H.~R.~B. Orlande}, \bibinfo{author}{M.~J. Colaco},
  \bibinfo{author}{O.~Fudym}, \bibinfo{author}{M.~El-Hafi},
  \bibinfo{author}{B.~Cuenot}, \bibinfo{author}{S.~Ricci},
\newblock \bibinfo{title}{{Application of particle filters to regional-scale
  wildfire spread}},
\newblock \bibinfo{journal}{{High Temperatures-High Pressures}}
  \bibinfo{volume}{43} (\bibinfo{year}{2014}) \bibinfo{pages}{p.415--440}.
  \URLprefix \url{https://hal.archives-ouvertes.fr/hal-01625026},
  \bibinfo{note}{european Conference on Thermophysical Properties
  (ECTP)European Conference on Thermophysical Properties (ECTP), Porto,
  PORTUGALPorto, PORTUGAL, SEP 05-05, 2014SEP 05-05, 2014}.
\bibitem[{Yoo and Wikle(2023)}]{yoo2023using}
\bibinfo{author}{M.~Yoo}, \bibinfo{author}{C.~K. Wikle},
\newblock \bibinfo{title}{Using echo state networks to inform physical models
  for fire front propagation},
\newblock \bibinfo{journal}{Spatial Statistics} \bibinfo{volume}{54}
  (\bibinfo{year}{2023}) \bibinfo{pages}{100732}. \URLprefix
  \url{https://www.sciencedirect.com/science/article/pii/S2211675323000076}.
  \DOIprefix\doi{https://doi.org/10.1016/j.spasta.2023.100732}.
\bibitem[{Baydin et~al.(2018)Baydin, Pearlmutter, Radul, and
  Siskind}]{baydin2018automatic}
\bibinfo{author}{A.~G. Baydin}, \bibinfo{author}{B.~A. Pearlmutter},
  \bibinfo{author}{A.~A. Radul}, \bibinfo{author}{J.~M. Siskind},
\newblock \bibinfo{title}{Automatic differentiation in machine learning: a
  survey},
\newblock \bibinfo{journal}{Journal of Machine Learning Research}
  \bibinfo{volume}{18} (\bibinfo{year}{2018}) \bibinfo{pages}{1--43}.
  \URLprefix \url{http://jmlr.org/papers/v18/17-468.html}.
\bibitem[{Karniadakis et~al.(2021)Karniadakis, Kevrekidis, Lu, Perdikaris,
  Wang, and Yang}]{Karniadakis2021Physics}
\bibinfo{author}{G.~E. Karniadakis}, \bibinfo{author}{I.~G. Kevrekidis},
  \bibinfo{author}{L.~Lu}, \bibinfo{author}{P.~Perdikaris},
  \bibinfo{author}{S.~Wang}, \bibinfo{author}{L.~Yang},
\newblock \bibinfo{title}{Physics-informed machine learning},
\newblock \bibinfo{journal}{Nature Reviews Physics} \bibinfo{volume}{3}
  (\bibinfo{year}{2021}) \bibinfo{pages}{422--440}. \URLprefix
  \url{https://doi.org/10.1038/s42254-021-00314-5}.
  \DOIprefix\doi{10.1038/s42254-021-00314-5}.
\bibitem[{Kollmannsberger et~al.(2021)Kollmannsberger, D'Angella, Jokeit, and
  Herrmann}]{Kollmannsberger2021Physics}
\bibinfo{author}{S.~Kollmannsberger}, \bibinfo{author}{D.~D'Angella},
  \bibinfo{author}{M.~Jokeit}, \bibinfo{author}{L.~Herrmann},
  \bibinfo{title}{Physics-Informed Neural Networks},
  \bibinfo{publisher}{Springer International Publishing},
  \bibinfo{address}{Cham}, \bibinfo{year}{2021}, pp. \bibinfo{pages}{55--84}.
  \URLprefix \url{https://doi.org/10.1007/978-3-030-76587-3_5}.
  \DOIprefix\doi{10.1007/978-3-030-76587-3_5}.
\bibitem[{H{\"u}llermeier and Waegeman(2021)}]{hullermeier2021aleatoric}
\bibinfo{author}{E.~H{\"u}llermeier}, \bibinfo{author}{W.~Waegeman},
\newblock \bibinfo{title}{Aleatoric and epistemic uncertainty in machine
  learning: An introduction to concepts and methods},
\newblock \bibinfo{journal}{Machine Learning} \bibinfo{volume}{110}
  (\bibinfo{year}{2021}) \bibinfo{pages}{457--506}.
\bibitem[{Sahli~Costabal et~al.(2020)Sahli~Costabal, Yang, Perdikaris, Hurtado,
  and Kuhl}]{Costabal2020Physics}
\bibinfo{author}{F.~Sahli~Costabal}, \bibinfo{author}{Y.~Yang},
  \bibinfo{author}{P.~Perdikaris}, \bibinfo{author}{D.~E. Hurtado},
  \bibinfo{author}{E.~Kuhl},
\newblock \bibinfo{title}{Physics-informed neural networks for cardiac
  activation mapping},
\newblock \bibinfo{journal}{Frontiers in Physics} \bibinfo{volume}{8}
  (\bibinfo{year}{2020}). \URLprefix
  \url{https://www.frontiersin.org/articles/10.3389/fphy.2020.00042}.
  \DOIprefix\doi{10.3389/fphy.2020.00042}.
\bibitem[{Zhang et~al.(2019)Zhang, Lu, Guo, and
  Karniadakis}]{zhang2019quantifying}
\bibinfo{author}{D.~Zhang}, \bibinfo{author}{L.~Lu}, \bibinfo{author}{L.~Guo},
  \bibinfo{author}{G.~E. Karniadakis},
\newblock \bibinfo{title}{Quantifying total uncertainty in physics-informed
  neural networks for solving forward and inverse stochastic problems},
\newblock \bibinfo{journal}{Journal of Computational Physics}
  \bibinfo{volume}{397} (\bibinfo{year}{2019}) \bibinfo{pages}{108850}.
  \URLprefix
  \url{https://www.sciencedirect.com/science/article/pii/S0021999119305340}.
  \DOIprefix\doi{https://doi.org/10.1016/j.jcp.2019.07.048}.
\bibitem[{Yang and Perdikaris(2019)}]{yang2019adversarial}
\bibinfo{author}{Y.~Yang}, \bibinfo{author}{P.~Perdikaris},
\newblock \bibinfo{title}{Adversarial uncertainty quantification in
  physics-informed neural networks},
\newblock \bibinfo{journal}{Journal of Computational Physics}
  \bibinfo{volume}{394} (\bibinfo{year}{2019}) \bibinfo{pages}{136--152}.
  \URLprefix
  \url{https://www.sciencedirect.com/science/article/pii/S0021999119303584}.
  \DOIprefix\doi{https://doi.org/10.1016/j.jcp.2019.05.027}.
\bibitem[{Gao and Ng(2022)}]{Gao2022Wasserstein}
\bibinfo{author}{Y.~Gao}, \bibinfo{author}{M.~K. Ng},
\newblock \bibinfo{title}{Wasserstein generative adversarial uncertainty
  quantification in physics-informed neural networks},
\newblock \bibinfo{journal}{Journal of Computational Physics}
  \bibinfo{volume}{463} (\bibinfo{year}{2022}) \bibinfo{pages}{111270}.
  \URLprefix
  \url{https://www.sciencedirect.com/science/article/pii/S0021999122003321}.
  \DOIprefix\doi{https://doi.org/10.1016/j.jcp.2022.111270}.
\bibitem[{Krishnapriyan et~al.(2021)Krishnapriyan, Gholami, Zhe, Kirby, and
  Mahoney}]{krishnapriyan2021characterizing}
\bibinfo{author}{A.~Krishnapriyan}, \bibinfo{author}{A.~Gholami},
  \bibinfo{author}{S.~Zhe}, \bibinfo{author}{R.~Kirby}, \bibinfo{author}{M.~W.
  Mahoney},
\newblock \bibinfo{title}{Characterizing possible failure modes in
  physics-informed neural networks},
\newblock in: \bibinfo{editor}{M.~Ranzato}, \bibinfo{editor}{A.~Beygelzimer},
  \bibinfo{editor}{Y.~Dauphin}, \bibinfo{editor}{P.~Liang},
  \bibinfo{editor}{J.~W. Vaughan} (Eds.), \bibinfo{booktitle}{Advances in
  Neural Information Processing Systems}, volume~\bibinfo{volume}{34},
  \bibinfo{publisher}{Curran Associates, Inc.}, \bibinfo{year}{2021}, pp.
  \bibinfo{pages}{26548--26560}. \URLprefix
  \url{https://proceedings.neurips.cc/paper/2021/file/df438e5206f31600e6ae4af72f2725f1-Paper.pdf}.
\bibitem[{Hornik et~al.(1989)Hornik, Stinchcombe, and
  White}]{hornik1989multilayer}
\bibinfo{author}{K.~Hornik}, \bibinfo{author}{M.~Stinchcombe},
  \bibinfo{author}{H.~White},
\newblock \bibinfo{title}{Multilayer feedforward networks are universal
  approximators},
\newblock \bibinfo{journal}{Neural Networks} \bibinfo{volume}{2}
  (\bibinfo{year}{1989}) \bibinfo{pages}{359--366}. \URLprefix
  \url{https://www.sciencedirect.com/science/article/pii/0893608089900208}.
  \DOIprefix\doi{https://doi.org/10.1016/0893-6080(89)90020-8}.
\bibitem[{Wang et~al.(2021{\natexlab{a}})Wang, Teng, and
  Perdikaris}]{wang2021understanding}
\bibinfo{author}{S.~Wang}, \bibinfo{author}{Y.~Teng},
  \bibinfo{author}{P.~Perdikaris},
\newblock \bibinfo{title}{Understanding and mitigating gradient flow
  pathologies in physics-informed neural networks},
\newblock \bibinfo{journal}{SIAM Journal on Scientific Computing}
  \bibinfo{volume}{43} (\bibinfo{year}{2021}{\natexlab{a}})
  \bibinfo{pages}{A3055--A3081}. \DOIprefix\doi{10.1137/20M1318043}.
\bibitem[{Wang et~al.(2021{\natexlab{b}})Wang, Wang, and
  Perdikaris}]{wang2021eigenvector}
\bibinfo{author}{S.~Wang}, \bibinfo{author}{H.~Wang},
  \bibinfo{author}{P.~Perdikaris},
\newblock \bibinfo{title}{On the eigenvector bias of fourier feature networks:
  From regression to solving multi-scale pdes with physics-informed neural
  networks},
\newblock \bibinfo{journal}{Computer Methods in Applied Mechanics and
  Engineering} \bibinfo{volume}{384} (\bibinfo{year}{2021}{\natexlab{b}})
  \bibinfo{pages}{113938}. \URLprefix
  \url{https://www.sciencedirect.com/science/article/pii/S0045782521002759}.
  \DOIprefix\doi{https://doi.org/10.1016/j.cma.2021.113938}.
\bibitem[{Wang et~al.(2022)Wang, Yu, and Perdikaris}]{wang2022when}
\bibinfo{author}{S.~Wang}, \bibinfo{author}{X.~Yu},
  \bibinfo{author}{P.~Perdikaris},
\newblock \bibinfo{title}{When and why pinns fail to train: A neural tangent
  kernel perspective},
\newblock \bibinfo{journal}{Journal of Computational Physics}
  \bibinfo{volume}{449} (\bibinfo{year}{2022}) \bibinfo{pages}{110768}.
  \URLprefix
  \url{https://www.sciencedirect.com/science/article/pii/S002199912100663X}.
  \DOIprefix\doi{https://doi.org/10.1016/j.jcp.2021.110768}.
\bibitem[{Mattey and Ghosh(2022)}]{mattey2022novel}
\bibinfo{author}{R.~Mattey}, \bibinfo{author}{S.~Ghosh},
\newblock \bibinfo{title}{A novel sequential method to train physics informed
  neural networks for allen cahn and cahn hilliard equations},
\newblock \bibinfo{journal}{Computer Methods in Applied Mechanics and
  Engineering} \bibinfo{volume}{390} (\bibinfo{year}{2022})
  \bibinfo{pages}{114474}. \URLprefix
  \url{https://www.sciencedirect.com/science/article/pii/S0045782521006939}.
  \DOIprefix\doi{https://doi.org/10.1016/j.cma.2021.114474}.
\bibitem[{Nascimento et~al.(2020)Nascimento, Fricke, and
  Viana}]{nascimento2020tutorial}
\bibinfo{author}{R.~G. Nascimento}, \bibinfo{author}{K.~Fricke},
  \bibinfo{author}{F.~A. Viana},
\newblock \bibinfo{title}{A tutorial on solving ordinary differential equations
  using python and hybrid physics-informed neural network},
\newblock \bibinfo{journal}{Engineering Applications of Artificial
  Intelligence} \bibinfo{volume}{96} (\bibinfo{year}{2020})
  \bibinfo{pages}{103996}. \URLprefix
  \url{https://www.sciencedirect.com/science/article/pii/S095219762030292X}.
  \DOIprefix\doi{https://doi.org/10.1016/j.engappai.2020.103996}.
\bibitem[{Yucesan and Viana(2020)}]{yucesan2020physics}
\bibinfo{author}{Y.~A. Yucesan}, \bibinfo{author}{F.~A. Viana},
\newblock \bibinfo{title}{A physics-informed neural network for wind turbine
  main bearing fatigue},
\newblock \bibinfo{journal}{International Journal of Prognostics and Health
  Management} \bibinfo{volume}{11} (\bibinfo{year}{2020}).
\bibitem[{Zubov et~al.(2021)Zubov, McCarthy, Ma, Calisto, Pagliarino, Azeglio,
  Bottero, Luj{\'a}n, Sulzer, Bharambe et~al.}]{zubov2021neuralpde}
\bibinfo{author}{K.~Zubov}, \bibinfo{author}{Z.~McCarthy},
  \bibinfo{author}{Y.~Ma}, \bibinfo{author}{F.~Calisto},
  \bibinfo{author}{V.~Pagliarino}, \bibinfo{author}{S.~Azeglio},
  \bibinfo{author}{L.~Bottero}, \bibinfo{author}{E.~Luj{\'a}n},
  \bibinfo{author}{V.~Sulzer}, \bibinfo{author}{A.~Bharambe}, et~al.,
\newblock \bibinfo{title}{Neuralpde: Automating physics-informed neural
  networks (pinns) with error approximations},
\newblock \bibinfo{journal}{arXiv preprint arXiv:2107.09443}
  (\bibinfo{year}{2021}).
\bibitem[{Bottero et~al.(2021)Bottero, Calisto, Graziano, Pagliarino, Scauda,
  Tiengo, and Azeglio}]{bottero2021physics}
\bibinfo{author}{L.~Bottero}, \bibinfo{author}{F.~Calisto},
  \bibinfo{author}{G.~Graziano}, \bibinfo{author}{V.~Pagliarino},
  \bibinfo{author}{M.~Scauda}, \bibinfo{author}{S.~Tiengo},
  \bibinfo{author}{S.~Azeglio}, \bibinfo{title}{Physics-Informed Machine
  Learning Simulator for Wildfire Propagation}, \bibinfo{type}{Technical
  Report}, Department of Physics, University of Turin, \bibinfo{year}{2021}.
\bibitem[{Rothermel(1972)}]{rothermel1972mathematical}
\bibinfo{author}{R.~C. Rothermel}, \bibinfo{title}{A mathematical model for
  predicting fire spread in wildland fuels}, volume \bibinfo{volume}{115},
  \bibinfo{publisher}{Intermountain Forest \& Range Experiment Station, Forest
  Service, US~…}, \bibinfo{year}{1972}.
\bibitem[{Green et~al.(1983)Green, Gill, and Noble}]{green1983fire}
\bibinfo{author}{D.~Green}, \bibinfo{author}{A.~Gill},
  \bibinfo{author}{I.~Noble},
\newblock \bibinfo{title}{Fire shapes and the adequacy of fire-spread models},
\newblock \bibinfo{journal}{Ecological Modelling} \bibinfo{volume}{20}
  (\bibinfo{year}{1983}) \bibinfo{pages}{33--45}. \URLprefix
  \url{https://www.sciencedirect.com/science/article/pii/0304380083900303}.
  \DOIprefix\doi{https://doi.org/10.1016/0304-3800(83)90030-3}.
\bibitem[{Sussman et~al.(1994)Sussman, Smereka, and Osher}]{sussman1994level}
\bibinfo{author}{M.~Sussman}, \bibinfo{author}{P.~Smereka},
  \bibinfo{author}{S.~Osher},
\newblock \bibinfo{title}{A level set approach for computing solutions to
  incompressible two-phase flow},
\newblock \bibinfo{journal}{Journal of Computational Physics}
  \bibinfo{volume}{114} (\bibinfo{year}{1994}) \bibinfo{pages}{146--159}.
  \URLprefix
  \url{https://www.sciencedirect.com/science/article/pii/S0021999184711557}.
  \DOIprefix\doi{https://doi.org/10.1006/jcph.1994.1155}.
\bibitem[{Raissi et~al.(2017)Raissi, Perdikaris, and
  Karniadakis}]{raissi2017physics}
\bibinfo{author}{M.~Raissi}, \bibinfo{author}{P.~Perdikaris},
  \bibinfo{author}{G.~E. Karniadakis},
\newblock \bibinfo{title}{Physics informed deep learning (part i): Data-driven
  solutions of nonlinear partial differential equations},
\newblock \bibinfo{journal}{arXiv preprint arXiv:1711.10561}
  (\bibinfo{year}{2017}).
\bibitem[{Goodfellow et~al.(2016)Goodfellow, Bengio, and
  Courville}]{goodfellow2016deep}
\bibinfo{author}{I.~Goodfellow}, \bibinfo{author}{Y.~Bengio},
  \bibinfo{author}{A.~Courville}, \bibinfo{title}{Deep learning},
  \bibinfo{publisher}{MIT press}, \bibinfo{year}{2016}.
\bibitem[{Rohrhofer et~al.(2021)Rohrhofer, Posch, and
  Geiger}]{rohrhofer2021pareto}
\bibinfo{author}{F.~M. Rohrhofer}, \bibinfo{author}{S.~Posch},
  \bibinfo{author}{B.~C. Geiger},
\newblock \bibinfo{title}{On the pareto front of physics-informed neural
  networks},
\newblock \bibinfo{journal}{arXiv preprint arXiv:2105.00862}
  (\bibinfo{year}{2021}).
\bibitem[{Schroeder et~al.(2014)Schroeder, Oliva, Giglio, and
  Csiszar}]{Schroeder2014New}
\bibinfo{author}{W.~Schroeder}, \bibinfo{author}{P.~Oliva},
  \bibinfo{author}{L.~Giglio}, \bibinfo{author}{I.~A. Csiszar},
\newblock \bibinfo{title}{The new viirs 375m active fire detection data
  product: Algorithm description and initial assessment},
\newblock \bibinfo{journal}{Remote Sensing of Environment}
  \bibinfo{volume}{143} (\bibinfo{year}{2014}) \bibinfo{pages}{85--96}.
  \URLprefix
  \url{https://www.sciencedirect.com/science/article/pii/S0034425713004483}.
  \DOIprefix\doi{https://doi.org/10.1016/j.rse.2013.12.008}.
\bibitem[{Blundell et~al.(2015)Blundell, Cornebise, Kavukcuoglu, and
  Wierstra}]{blundell2015weight}
\bibinfo{author}{C.~Blundell}, \bibinfo{author}{J.~Cornebise},
  \bibinfo{author}{K.~Kavukcuoglu}, \bibinfo{author}{D.~Wierstra},
\newblock \bibinfo{title}{Weight uncertainty in neural network},
\newblock in: \bibinfo{booktitle}{International conference on machine
  learning}, \bibinfo{organization}{PMLR}, \bibinfo{year}{2015}, pp.
  \bibinfo{pages}{1613--1622}.
\bibitem[{Goan and Fookes(2020)}]{goan2020bayesian}
\bibinfo{author}{E.~Goan}, \bibinfo{author}{C.~Fookes},
\newblock \bibinfo{title}{Bayesian neural networks: An introduction and
  survey},
\newblock \bibinfo{journal}{Case Studies in Applied Bayesian Data Science: CIRM
  Jean-Morlet Chair, Fall 2018}  (\bibinfo{year}{2020})
  \bibinfo{pages}{45--87}.
\bibitem[{Blei et~al.(2017)Blei, Kucukelbir, and
  McAuliffe}]{blei2017variational}
\bibinfo{author}{D.~M. Blei}, \bibinfo{author}{A.~Kucukelbir},
  \bibinfo{author}{J.~D. McAuliffe},
\newblock \bibinfo{title}{Variational inference: A review for statisticians},
\newblock \bibinfo{journal}{Journal of the American statistical Association}
  \bibinfo{volume}{112} (\bibinfo{year}{2017}) \bibinfo{pages}{859--877}.
\bibitem[{Kingma and Welling(2014)}]{kingma2014autoencoding}
\bibinfo{author}{D.~P. Kingma}, \bibinfo{author}{M.~Welling},
\newblock \bibinfo{title}{Auto-encoding variational bayes},
\newblock in: \bibinfo{booktitle}{Proceedings of the Second International
  Conference on Learning Representations (ICLR 2014)}, \bibinfo{year}{2014}.
\bibitem[{Kingma and Ba(2014)}]{kingma2014adam}
\bibinfo{author}{D.~P. Kingma}, \bibinfo{author}{J.~Ba},
\newblock \bibinfo{title}{Adam: A method for stochastic optimization},
\newblock \bibinfo{journal}{arXiv preprint arXiv:1412.6980}
  (\bibinfo{year}{2014}).
\bibitem[{Sullivan et~al.(2018)Sullivan, Cruz, Hilton, Plucinski, and
  Hurley}]{sullivan2018study}
\bibinfo{author}{A.~L. Sullivan}, \bibinfo{author}{M.~G. Cruz},
  \bibinfo{author}{J.~E. Hilton}, \bibinfo{author}{M.~P. Plucinski},
  \bibinfo{author}{R.~Hurley}, \bibinfo{title}{Study of growth of free-burning
  grass fires from point ignition}, \bibinfo{publisher}{Imprensa da
  Universidade de Coimbra}, \bibinfo{address}{Coimbra}, \bibinfo{year}{2018},
  pp. \bibinfo{pages}{643--649}.
  \DOIprefix\doi{https://doi.org/10.14195/978-989-26-16-506_71}.
\bibitem[{Neal et~al.(2011)}]{neal2011mcmc}
\bibinfo{author}{R.~M. Neal}, et~al.,
\newblock \bibinfo{title}{Mcmc using hamiltonian dynamics},
\newblock \bibinfo{journal}{Handbook of markov chain monte carlo}
  \bibinfo{volume}{2} (\bibinfo{year}{2011}) \bibinfo{pages}{2}.
\bibitem[{McDermott and Wikle(2019)}]{mcdermott2019bayesian}
\bibinfo{author}{P.~L. McDermott}, \bibinfo{author}{C.~K. Wikle},
\newblock \bibinfo{title}{Bayesian recurrent neural network models for
  forecasting and quantifying uncertainty in spatial-temporal data},
\newblock \bibinfo{journal}{Entropy} \bibinfo{volume}{21}
  (\bibinfo{year}{2019}) \bibinfo{pages}{184}.
\bibitem[{Pausas and Keeley(2021)}]{pausas2021wildfires}
\bibinfo{author}{J.~G. Pausas}, \bibinfo{author}{J.~E. Keeley},
\newblock \bibinfo{title}{Wildfires and global change},
\newblock \bibinfo{journal}{Frontiers in Ecology and the Environment}
  \bibinfo{volume}{19} (\bibinfo{year}{2021}) \bibinfo{pages}{387--395}.
\bibitem[{Hilton et~al.(2018)Hilton, Sullivan, Swedosh, Sharples, and
  Thomas}]{hilton2018incorporating}
\bibinfo{author}{J.~Hilton}, \bibinfo{author}{A.~Sullivan},
  \bibinfo{author}{W.~Swedosh}, \bibinfo{author}{J.~Sharples},
  \bibinfo{author}{C.~Thomas},
\newblock \bibinfo{title}{Incorporating convective feedback in wildfire
  simulations using pyrogenic potential},
\newblock \bibinfo{journal}{Environmental Modelling \& Software}
  \bibinfo{volume}{107} (\bibinfo{year}{2018}) \bibinfo{pages}{12--24}.
  \URLprefix
  \url{https://www.sciencedirect.com/science/article/pii/S1364815217309593}.
  \DOIprefix\doi{https://doi.org/10.1016/j.envsoft.2018.05.009}.

\end{thebibliography}


%
%
%


\end{document}